\documentclass[journal]{IEEEtran}

\usepackage{graphicx} \graphicspath{ {figures/} }
\usepackage{amsmath,amssymb,mathabx}
\usepackage{algorithmic}
\usepackage[linesnumbered,ruled,vlined]{algorithm2e}
\usepackage{acronym}
\usepackage{enumitem}
\usepackage{booktabs}
\usepackage{hyperref}
\usepackage{balance}
\usepackage{xspace,setspace}
\usepackage[skip=3pt,font=small]{subcaption}
\usepackage[skip=3pt,font=small]{caption}
\usepackage[dvipsnames]{xcolor}
\usepackage[capitalise]{cleveref}
\usepackage{tabularx,colortbl,multirow,array,makecell}
\usepackage{overpic}
\usepackage{cite}
\usepackage{tikz}

\makeatletter
\DeclareRobustCommand\onedot{\futurelet\@let@token\@onedot}
\def\@onedot{\ifx\@let@token.\else.\null\fi\xspace}
\def\eg{\emph{e.g}\onedot} 
\def\ie{\emph{i.e}\onedot}

\makeatother

\makeatletter
\def\BState{\State\hskip-\ALG@thistlm}
\makeatother

\makeatletter
\renewcommand{\paragraph}{%
  \@startsection{paragraph}{4}%
  {\z@}{0ex \@plus 0ex \@minus 0ex}{-1em}%
  {\hskip\parindent\normalfont\normalsize\bfseries}%
}
\makeatother

\crefname{algocf}{alg.}{algs.}
\Crefname{algocf}{Algorithm}{Algorithms}
\crefname{section}{Sec.}{Secs.}
\Crefname{section}{Section}{Sections}

\definecolor{gblue}{HTML}{4285F4}
\definecolor{gred}{HTML}{DB4437}

\acrodef{dof}[DoF]{Degree of Freedom}
\acrodef{tamp}[TAMP]{Task and Motion Planning}
\acrodef{ik}[IK]{Inverse Kinematics}
\acrodef{urdf}[URDF]{Universal Robot Description Format}

\newcommand{\name}{\textsc{IKDiffuser}\xspace}

\let\oldnl\nl
\newcommand{\nosemic}{\SetEndCharOfAlgoLine{\relax}}
\newcommand{\nonl}{\renewcommand{\nl}{\let\nl\oldnl}}

\newcommand\blfootnote[1]{%
  \begingroup
  \renewcommand\thefootnote{}\footnote{#1}%
  \addtocounter{footnote}{-1}%
  \endgroup
}

\title{\textsc{IKDiffuser}: a Diffusion-based Generative \\ Inverse Kinematics Solver for Kinematic Trees}

\author{
    Zeyu~Zhang, 
    Ziyuan~Jiao
}

\begin{document}


\twocolumn[{
\renewcommand\twocolumn[1][]{#1}
\maketitle
\thispagestyle{empty}
\pagestyle{empty}
\vspace{-0.5cm}
\begin{center}
    \centering
    \captionsetup{type=figure}
        \includegraphics[width=\linewidth]{figures/teaser/teaser_row_1.pdf}
        \caption{\textbf{\name: A generative \ac{ik} solver for kinematic trees.} 
        The solver is capable of generating \ac{ik} solutions for robots with arbitrary kinematic trees: 
        (a)~\textit{Single serial chain}---Franka Research 3; 
        (b)~\textit{Dual serial chains with fixed base}---Baxter dualarm; 
        (c)~\textit{Dual serial chains with shared joints}---Dual RealMan-75 with waist; 
        (d)~\textit{Dual serial chains with a mobile base}---Mobile dual UR5e; 
        (e)~\textit{Humanoid with four serial limbs}---NASA Robonaut and Unitree G1; 
        (f)~\textit{Four-fingered dexterous hand with a floating base}---Leap Hand.}
    \label{fig:teaser}
    \vspace{8pt}
\end{center}
}]

\blfootnote{Zeyu Zhang and Ziyuao Jiao are with State Key Laboratory of General Artificial Intelligence, Beijing Institute for General Artificial Intelligence (BIGAI), Beijing 100080, China (emails: \{\texttt{zhangzeyu},\texttt{jiaoziyuan}\}@bigai.ai).}

\begin{abstract}
Solving Inverse Kinematics (IK) for arbitrary kinematic trees presents significant challenges due to their high-dimensionality, redundancy, and complex inter-branch constraints. 
Conventional optimization-based solvers can be sensitive to initialization and suffer from local minima or conflicting gradients. At the same time, existing learning-based approaches are often tied to a predefined number of end-effectors and a fixed training objective, limiting their reusability across various robot morphologies and task requirements.
To address these limitations, we introduce \textsc{IKDiffuser}, a scalable IK solver built upon conditional diffusion-based generative models, which learns the distribution of the configuration space conditioned on end-effector poses.
We propose a structure-agnostic formulation that represents end-effector poses as a sequence of tokens, leading to a unified framework that handles varying numbers of end-effectors while learning the implicit kinematic structures entirely from data.
Beyond standard \ac{ik} generation, \name handles partially specified goals via a masked marginalization mechanism that conditions only on a subset of end-effector constraints. 
Furthermore, it supports adding task objectives at inference through objective-guided sampling, enabling capabilities such as warm-start initialization and manipulability maximization without retraining.
Extensive evaluations across seven diverse robotic platforms demonstrate that \textsc{IKDiffuser} significantly outperforms state-of-the-art baselines in accuracy, solution diversity, and collision avoidance. 
Moreover, when used to initialize optimization-based solvers, \textsc{IKDiffuser} significantly boosts success rates on challenging redundant systems with high Degrees of Freedom (DoF), such as the 29-DoF Unitree G1 humanoid, from 21.01\% to 96.96\% while reducing computation time to the millisecond range.
Overall, \textsc{IKDiffuser} offers a unified, scalable, and robust solution for the IK problem in robotic systems with complex kinematics structures.
\end{abstract}



\section{Introduction}\label{sec:introduction}

\setstretch{1.005}

\IEEEPARstart{I}{nverse} Kinematics (IK) is a fundamental problem in robotics, where the goal is to compute joint configurations that achieve a desired end-effector pose in operational space~\cite{siciliano2009robotics}. This capability underlies a wide range of applications, from achieving desired grasping poses in cluttered scenes~\cite{jiao2021consolidated,jiao2025integration} to ensuring precise tool alignment in manufacturing operations~\cite{chiaverini1994review}. Traditional solvers have proven highly effective for single fixed-base manipulators, where the robot exhibits a serial kinematic chain with limited redundancy~\cite{beeson2015trac,carpentier2019pinocchio}. However, modern robotic platforms are increasingly built upon tree-like kinematic structures (\ie, kinematic trees with multiple ends, redundant and coupled joints)---such as multi-arm mobile robots~\cite{fu2024mobile}, humanoids~\cite{tevatia2000inverse}, and dexterous hands~\cite{shaw2023leap}---where multiple chains interact through shared roots, coupled joints, redundant DoFs, and self-collisions. For these systems, the IK problem is significantly more challenging.

The difficulty of solving IK for kinematic trees stems from the higher configuration-space dimensionality, stronger inter-branch coupling~\cite{wang2016whole}, and more complex self-collision constraints that fragment the feasible configuration subspaces~\cite{dai2019global}. 
Among existing IK solvers, optimization-based solvers are highly sensitive to initialization. They can fall in poor local minima when branch-wise gradients conflict~\cite{kim2025pyroki}, while gradient-free methods remain too computationally expensive~\cite{wu2021t}. 
In recent years, learning-based approaches have been explored~\cite{ames2022ikflow,chi2023diffusion,limoyo2024generative}, yet most are tightly coupled to a specific robot morphology and training objective: they assume a predefined number of end-effectors and kinematic topologies, making it difficult to scale to arbitrary kinematic structures without redesigning the architecture. 
Moreover, these methods rarely support adding task-specific objectives without retraining the entire model.
As a result, separate learned models must be obtained across different end-effector configurations and objective sets---even for the same robot---which undermines reusability and scalability. 
These limitations motivate a unified, structure-agnostic approach that generalizes across kinematic trees, while preserving high-precision solutions and enabling customization of task-specific objectives at inference time.

In this paper, we introduce \name, a diffusion-based generative \ac{ik} solver for kinematic trees. Unlike prior learning-based \ac{ik} methods that target single serial manipulators or directly regress joint states, \name models a distribution over the full configuration space of a kinematic tree, capturing redundancy and inter-branch constraints within a unified formulation.
Rather than concatenating all target end-effector poses into a fixed-length vector, we represent them as a \emph{sequence} of conditioning tokens.
This sequential formulation is \emph{structure-agnostic}: it does not hard-code the kinematic structure, but instead allows the attention mechanism to implicitly infer inter-branch dependencies and constraints entirely from data, regardless of the number of end-effectors.
Since end-effector poses are provided token-wise, \name also supports partially specified goals by conditioning only on available tokens for a subset of end-effectors. To strengthen this capability, we introduce a masked marginalization strategy during training that enforces constraints on specified end-effectors while treating the remaining ones as latent.
Moreover, \name supports the integration of task-specific objectives, such as manipulability maximization and warm-start initialization, through guided sampling at inference time, without requiring any retraining or finetuning of the model.

We evaluate \name on seven robotic platforms spanning diverse kinematic structures, including standard single fixed-base manipulators, (mobile) dual-arm robots, humanoids, and dexterous hands (see \cref{fig:teaser}). Across all cases, \name achieves the highest accuracy and solution diversity compared to the state-of-the-art baselines. 
When used to seed an optimization-based \ac{ik} solver, \name can boost the success rate of finding feasible solutions (within $10^{-3}$~mm and $10^{-3}$~deg precision) on the 29-DoF Unitree G1 humanoid from 21.01\% to 96.96\%, while reducing the time to obtain 128 feasible solutions from 1830~ms to 280~ms.
Furthermore, through task-specific and marginal inference, \name supports manipulability maximization, warm-start initialization, and partially constrained end-effector poses---all without retraining the model. These results underscore the scalability and practicality of our diffusion-based generative IK solver for robotic systems
that are evolving toward greater kinematic complexity.

\setstretch{1.}

This work makes the following contributions:
\begin{enumerate}[label=\arabic*), leftmargin=*, noitemsep, nolistsep]
    \item We present \name, a diffusion-based generative IK solver that produces fast and diverse solutions for robots with arbitrary kinematic trees. \name also enables optimization-based solvers to achieve high-precision solutions significantly faster by seeding them.
    
    \item We propose a structure-agnostic conditioning formulation that represents end-effector goals as a sequence of conditioning tokens, and design a network architecture that can handle varying numbers of end-effectors within arbitrary kinematic trees within a unified formulation.
    
    \item We introduce a guided sampling strategy that incorporates task-specific objectives to shape the solution space, allowing \name to generate desired solutions at inference without retraining the model.
    
    \item We employ a masked marginalization strategy during training, enabling \name to enforce only a specified number of end-effector poses, thereby improving applicability to tasks with partial constraints.
    
    \item We provide extensive benchmarking against optimization- and learning-based baselines, demonstrating consistent improvements in precision, efficiency, and adaptability across seven robotic platforms.

    \item We develop a toolchain that efficiently collects data for new robots with arbitrary kinematic tree structures and automatically trains an IK solver with \name. The codebase will be released publicly to support reproducibility and further customized development.
\end{enumerate}

\subsection{Overview}
The remainder of this paper is organized as follows. \cref{sec:relatedwork} reviews the existing literature on inverse kinematics, contrasting traditional analytical and optimization-based methods with emerging learning-based and generative approaches. \cref{sec:method} details the methodology of \name, formulating the IK problem for kinematic trees as a probabilistic generative task. It introduces the structure-agnostic representation, followed by the algorithms for masked marginalization and objective-guided sampling. \cref{sec:experiment} presents experimental evaluations across robotic platforms with diverse kinematic structures, analyzing solution quality, computational efficiency, and the effectiveness of using \name to seed optimization solvers. Finally, \cref{sec:discussion} discusses key findings, limitations, and future directions, followed by a conclusion in \cref{sec:conclusion}.

\setstretch{1.02}

\section{Related Work}\label{sec:relatedwork}
\textbf{Solving IK problems for single chains:} Research on \ac{ik} began with single fixed-base manipulators, where the problem is comparatively well understood. 
Early analytical formulations exploited geometric and trigonometric properties to derive closed-form solutions~\cite{siciliano2009robotics}. 
These methods are computationally efficient but inherently limited to robots with simple structures, such as 6-DoF arms, and do not scale well to redundant systems. 
In such systems, the IK solution is generally non-unique, admitting infinitely many valid joint configurations for the same end-effector pose, which necessitates redundancy resolution to select among candidate solutions. 
Numerical approaches, such as Jacobian-based methods~\cite{wolovich1984computational,wampler1986manipulator,beeson2015trac,carpentier2019pinocchio} and optimization-based variants~\cite{wang1991combined,zhao1994inverse,kim2025pyroki}, offer greater flexibility by iteratively solving constrained optimization problems. 
However, their convergence strongly depends on initialization, and they are prone to local minima. 
While suitable for single arms in low-dimensional spaces, these methods often struggle to provide robust or diverse solutions for high-dimensional kinematic trees.

\textbf{Solving IK problems for kinematic trees:} 
Extending existing IK solvers to kinematic trees introduces significant complexity. 
Unlike single serial chains, these systems require simultaneous coordination across multiple branches, where shared joints influence the reachability of several end-effectors. 
This coupling significantly increases the search space complexity, a challenge further exacerbated by conflicting gradients arising from joint limits, inter-branch constraints, and task-specific requirements~\cite{dai2019global}. 
Classical optimization-based solvers struggle under these conditions, often necessitating human-crafted objectives or constraints and yielding solutions that lack both robustness and diversity~\cite{kim2025pyroki}. 
To address the limitations of gradient-based solvers, gradient-free approaches such as genetic algorithms and evolutionary strategies have been applied to IK~\cite{lopez2018inverse,wu2021t,sundaralingam2023curobo}.
While these methods can effectively explore complex and discontinuous search spaces without being trapped by local minima, they remain computationally expensive due to the cost of evaluating large populations. 
Consequently, their inconsistency and high runtime costs make them unsuitable for real-time deployment, where rapid, reliable IK generation is critical.

\textbf{Learning-based IK solvers:} Learning-based IK solvers have been introduced as a way to bypass iterative optimization by directly mapping desired end-effector poses in task space to joint configurations~\cite{oyama2001inverse,d2001learning,kim2021learning,ames2022ikflow,ardizzoneanalyzing,limoyo2024generative}.
While these methods can generate multiple candidate solutions via sampling and often improve speed and diversity, several limitations remain.
First, a direct mapping from task space (end-effector poses) to configuration space (joint state) typically assumes a fixed kinematic topology and a fixed number of end-effectors.
Since their formulation does not explicitly represent the inter-branch constraints, it is difficult to support varying branch counts, coupled joints, or end-effector sets without redesigning the model architecture and retraining from scratch.
A notable exception is GGIK~\cite{limoyo2024generative}, which employs a graph-based CVAE to generalize across morphologies by inferring intermediate Euclidean geometry. However, this reliance on an indirect geometric representation often leads to degraded solution precision when applied to unseen robots.
Second, these methods often scale poorly with dimensionality, since the solution manifolds of high-DoF or heavily branched systems become increasingly complex.
Third, most models are trained for a specific objective and provide limited inference-time adaptability. Incorporating task preferences, such as warm-start proximity, manipulability, or partially constrained end-effector goals, typically requires retraining the entire model.

\textbf{Diffusion-based robot motion generation:} Diffusion models and other generative paradigms have recently gained traction in robotics, particularly for trajectory generation and policy learning~\cite{chi2023diffusion,yan2025m,lin2025pp}. 
These approaches effectively capture complex motion distributions and scale favorably to high-dimensional spaces. 
However, these methods usually target task-level motion generation conditioned on task specifications and sensory observations, rather than producing explicit \ac{ik} solutions under hard geometric and kinematic constraints.
As a result, they generalize poorly to unseen task--scene combinations, and adapting them often requires changing the conditioning interface and collecting additional data.
In many applications---such as assembly tasks with tight tolerances~\cite{wang2019inverse}, motion planning in constrained environments~\cite{morgan2024cppflow}, or cross-embodiment applications~\cite{lu2024mobile,chi2024universal}---an explicit and accurate \ac{ik} solver remains indispensable.
This motivates a diffusion-based \ac{ik} solver as a reusable primitive: by conditioning directly on end-effector goals rather than task- or scene-specific context, it can complement downstream planners and optimization-based solvers in real-world deployment.

\textbf{Positioning of \name:} Our work builds on these lines of research by introducing \name, a diffusion-based generative \ac{ik} solver for arbitrary kinematic trees. 
Rather than relying on a fixed-length pose encoding and directly regressing joint configurations, \name models a distribution over the configuration space and conditions on end-effector goals as a sequence of tokens. This yields a structure-agnostic formulation that learns directly from data, bypassing intermediate morphological representations and eliminating the need to redesign the model architecture for different robots.
Moreover, \name supports partially specified goals through masked marginalization and incorporates task-specific objectives (\eg, warm-start initialization and manipulability maximization) via guided sampling at inference time. This enables flexible adaptation to new tasks and constraints without retraining: a capability that distinguishes our approach from previous learning-based solvers.
Importantly, \name serves as a complement to, rather than a replacement for, optimization-based solvers. By seeding these solvers with diverse \ac{ik} candidates, we combine the strengths of both paradigms: the resulting framework is fast, diverse, precise, and adaptable, making it highly suitable for practical applications in robotic systems with complex kinematic structures.



\section{IKDiffuser}\label{sec:method}

This section introduces \name, a unified, structure-agnostic generative framework designed to solve the \ac{ik} problem for robots with arbitrary kinematic trees. 
In the following, we first introduce the mathematical preliminaries of the IK problem and diffusion models. We then detail our structure-agnostic formulation, the training objective, and the inference strategies that enable task-specific guidance and the handling of partial end-effector constraints.

\subsection{Preliminary}\label{sec:preliminary}

\textbf{Problem Definition: }
Consider a robotic system with $N$ end-effectors. The objective is to determine the optimal joint configuration $\boldsymbol{q} \in \mathcal{C}$ that satisfies the sequence of desired end-effector poses $\mathcal{X} = \langle \boldsymbol{x}_1, \boldsymbol{x}_2, ..., \boldsymbol{x}_N \rangle$, where $\boldsymbol{x}_i \in SE(3)$. The pose of the $i$-th end-effector is defined by the forward kinematics function $f_i: \mathcal{C} \mapsto SE(3)$, such that $\boldsymbol{x}_i = f_i(\boldsymbol{q})$. Here, $f_i$ is governed by the serial kinematic chain $\textbf{K}_i$ of the $i$-th arm. Consequently, the \ac{ik} problem is formulated as the optimization:
\begin{equation}
    \boldsymbol{q}^* =\min_{\boldsymbol{q}}~\text{Cost}(\boldsymbol{q}, \mathcal{X}),
\end{equation}
where $\text{Cost}(\cdot)$ represents the cost function. 
In practice, this cost function typically quantifies the deviation between the desired poses $\mathcal{X}$ and the actual poses derived from $\boldsymbol{q}$, while also incorporating task-specific objectives to accommodate customized requirements.

\textbf{Diffusion Model:}
Diffusion models~\cite{ho2020denoising,song2019generative} are generative processes that synthesize data from a target distribution via iterative denoising. 
The process comprises two phases: forward and reverse diffusion. 
The forward process, defined by $\mathcal{N}(\boldsymbol{q}^t; \sqrt{1 - \beta^t}\boldsymbol{q}^{t-1}, \beta^t \boldsymbol{I})$, progressively corrupts the data sample $\boldsymbol{q}^0$ into an isotropic Gaussian distribution over $T$ steps, where $\beta^t \in \mathbb{R}$ controls the noise variance. 
Conversely, the reverse process employs a neural network $p_\theta(\boldsymbol{q}^{t-1} | \boldsymbol{q}^t)$ to recover the original sample $\boldsymbol{q}^0$ by approximating the denoising score across the noise scales observed during the forward phase. 
In this work, we propose a probabilistic formulation that leverages the diffusion process to model the joint distribution over the configuration space $\mathcal{C}$ and the operational space ${SE(3)}^{\otimes N}$, enabling the efficient generation of diverse \ac{ik} solutions from Gaussian noise.

\setstretch{1.}

\subsection{Problem Formulation}
\name formulates the \ac{ik} problem as probabilistic inference via \textit{analysis by synthesis}, where the model repeatedly ``synthesizes'' candidate joint configurations and evaluates how well they explain the desired end-effector constraints and objectives. 
This perspective yields diverse solutions by sampling from a learned latent distribution over the joint configuration space $\mathcal{C}$, conditioned on the desired end-effector poses $\mathcal{X}$ and task-specific objectives $\mathcal{J}$.
Crucially, instead of encoding the desired poses as a fixed-length vector, we represent them as \emph{a sequence}, allowing the model to capture the joint distribution over a variable number of end-effectors and thereby learn the inter-branch constraints induced by arbitrary kinematic trees.
We decompose the posterior distribution as:
\begin{align}
\begin{split}
    p(\boldsymbol{q}^0 | \mathcal{X}, \mathcal{J})=~
    & \frac{p_{\theta}(\boldsymbol{q}^0 | \mathcal{X}) p_{\varphi}(\mathcal{J} | \boldsymbol{q}^0, \mathcal{X})}{p(\mathcal{J} | \mathcal{X})} \\
    \propto~&
    \underbrace{p_{\theta}(\boldsymbol{q}^0 | \mathcal{X})}_{%
        \substack{\text{inverse}\\\text{kinematics}}
    }
    \;
    \underbrace{p_{\varphi}(\mathcal{J} | \boldsymbol{q}^0, \mathcal{X})}_{%
        \substack{\text{task-specific}\\\text{objectives}}
    } .
\end{split}
\end{align}
Here, $\boldsymbol{q}^0$ denotes the final denoised joint configuration at timestep $t=0$ of the diffusion process. The term $p_{\theta}(\boldsymbol{q}^0 | \mathcal{X})$ captures the solution space of the inverse kinematics problem, i.e., the distribution of feasible joint configurations given the end-effector goals, while $p_{\varphi}(\mathcal{J} | \boldsymbol{q}^0, \mathcal{X})$ models the likelihood that a configuration $\boldsymbol{q}$ satisfies the task-specific objectives $\mathcal{J}$.
In practice, $\mathcal{J}$ collects differentiable scalar or vector-valued objectives, and we convert them into a probability-like term via an energy-based construction (see \cref{subsec:objective} for details).
Thus, $p_{\varphi}$ is interpreted as a surrogate likelihood induced by these objectives rather than as a generative model of $\mathcal{J}$ itself.
Modeling objectives in this way enables inference-guided sampling during the diffusion process; consequently, finding \ac{ik} solutions reduces to sampling from the learned posterior $p(\boldsymbol{q}^0 | \mathcal{X}, \mathcal{J})$.

\textbf{Inverse Kinematics: }
We model the feasible solution space $p_{\theta}(\boldsymbol{q}^0 | \mathcal{X})$ using a conditional diffusion model~\cite{ho2020denoising, song2019generative}. 
This model learns to iteratively denoise Gaussian noise into a valid joint configuration through the following reverse process:
\begin{align}
    p_{\theta}(\boldsymbol{q}^0 | \mathcal{X}) & = \int p(\boldsymbol{q}^T) \prod_{t=1}^{T} p(\boldsymbol{q}^{t-1} | \boldsymbol{q}^t, \mathcal{X})~\text{d}\boldsymbol{q}^{1:T} \\
    p_\theta(\boldsymbol{q}^{t-1} | \boldsymbol{q}^t, \mathcal{X}) & = \mathcal{N}\left(\boldsymbol{q}^{t-1}; \boldsymbol{\mu}_\theta(\boldsymbol{q}^t, \mathcal{X}, t), \Sigma_\theta(\boldsymbol{q}^t, \mathcal{X}, t)\right),
    \label{eqn:rev_diffusion}
\end{align}
where $\boldsymbol{q}^T \sim \mathcal{N}(\boldsymbol{0}, \boldsymbol{I})$ is sampled from a standard normal prior, and \cref{eqn:rev_diffusion} defines the reverse transition parameterized by a neural network predicting $\boldsymbol{\mu}_\theta$ and $\Sigma_\theta$.

\textbf{Task-specific Objectives:} The term $p_{\varphi}(\mathcal{J} | \boldsymbol{q}^0, \mathcal{X})$ introduces auxiliary constraints over the sampled configurations $\boldsymbol{q}^0$ relative to the end-effector goals $\mathcal{X}$.
These objectives can be explicitly incorporated into the diffusion process to guide the sampling process towards the objective $\mathcal{J}$ without the need to retrain the underlying diffusion model $p_\theta(\boldsymbol{q}^{t-1} | \boldsymbol{q}^t, \mathcal{X})$~\cite{dhariwal2021diffusion}.
This flexibility allows for customized applications of diverse, differentiable task requirements.
We detail the specific objectives implementation in \name in \cref{subsec:objective}.

\subsection{Model Learning}

As shown in \cref{eqn:rev_diffusion}, the reverse diffusion process $ p_\theta(\boldsymbol{q}^{t-1} | \boldsymbol{q}^t, \mathcal{X})$ is modeled as a Gaussian distribution with mean $\boldsymbol{\mu}_\theta$ and covariance $\Sigma_\theta$.
To derive these parameters, we first recall the forward noising process with a fixed variance schedule $\{\beta^t\}_{t=1}^T$:
\begin{align}
    p(\boldsymbol{q}^t | \boldsymbol{q}^{t-1}, \mathcal{X}) &= \mathcal{N}\left(\boldsymbol{q}^t; \sqrt{1 - \beta^t}\boldsymbol{q}^{t-1}, \beta^t \boldsymbol{I}\right),
    \label{eqn:forward_diffusion}
\end{align}
which yields the closed-form distribution of $\boldsymbol{q}^t$ conditioned on the initial configuration $\boldsymbol{q}^0$:
\begin{align}
    p(\boldsymbol{q}^t | \boldsymbol{q}^0, \mathcal{X}) &= \mathcal{N}\left(\boldsymbol{q}^t;  \sqrt{\bar{\alpha}^t}\boldsymbol{q}^0, (1-\bar{\alpha}^t)\mathbf{I}\right),
    \label{eqn:forward_diffusion_nice}
\end{align}
where $\alpha^t = 1-\beta^t$ and $\bar{\alpha}^t = \prod_{i=1}^t \alpha^i$.
By Bayes' rule, the reverse conditional probability becomes tractable when conditioned on $\boldsymbol{q}^0$:
\begin{align}
    p(\boldsymbol{q}^{t-1} | \boldsymbol{q}^t, \boldsymbol{q}^0, \mathcal{X})
     = &~\frac{p(\boldsymbol{q}^t | \boldsymbol{q}^{t-1}, \boldsymbol{q}^0, \mathcal{X}) p(\boldsymbol{q}^{t-1} | \boldsymbol{q}^0, \mathcal{X})}{p(\boldsymbol{q}^t | \boldsymbol{q}^0, \mathcal{X})} \label{eqn:bayes_reverse_1} \\
    \propto &~p(\boldsymbol{q}^t | \boldsymbol{q}^{t-1}, \mathcal{X}) p(\boldsymbol{q}^{t-1} | \boldsymbol{q}^0, \mathcal{X}), \label{eqn:bayes_reverse_2}
\end{align}
where the second equality follows from the conditional independence of $\boldsymbol{q}^t$ and $\boldsymbol{q}^0$ given $\boldsymbol{q}^{t-1}$.

Since both terms in \cref{eqn:bayes_reverse_2} are Gaussian, the reverse denoising transition $p(\boldsymbol{q}^{t-1} | \boldsymbol{q}^t, \boldsymbol{q}^0, \mathcal{X})$ is also Gaussian.
Substituting \cref{eqn:forward_diffusion} and \cref{eqn:forward_diffusion_nice} into \cref{eqn:bayes_reverse_2}, we obtain the closed-form reverse denoising transition with mean $\tilde{\boldsymbol{\mu}}_t(\boldsymbol{q}^t,\boldsymbol{q}^0, \mathcal{X})$ and covariance $\tilde{\Sigma}^t$:
\begin{align}
    \tilde{\boldsymbol{\mu}}^t(\boldsymbol{q}^t,\boldsymbol{q}^0, \mathcal{X}) &= \frac{1}{\sqrt{\alpha^t}}\left(\boldsymbol{q}^t - \frac{1-\alpha^t}{1-\bar{\alpha}^t}\bigl(\boldsymbol{q}^t - \sqrt{\bar{\alpha}^t}\boldsymbol{q}^0\bigr)\right), 
    \label{eqn:reverse_mu_t}\\
    \tilde{\Sigma}^t &= \frac{1-\bar{\alpha}_{t-1}}{1-\bar{\alpha}_t}\beta^t \mathbf{I}.
    \label{eqn:reverse_sigma_t}
\end{align}

In practice, $\boldsymbol{q}^0$ is not available at inference time, preventing direct computation of $\tilde{\boldsymbol{\mu}}^t$ and $\tilde{\Sigma}^t$.
Rather than directly learning $\boldsymbol{\mu}_\theta$ and $\Sigma_\theta$ to estimate the mean and covariance of the reverse transition $p(\boldsymbol{q}^{t-1} | \boldsymbol{q}^t, \boldsymbol{q}^0, \mathcal{X})$, we follow the standard approach of training a noise prediction model~\cite{ho2020denoising} to estimate the noise $\epsilon^t$ added at timestep $t$, and use this estimate to parameterize the analytic expressions in \cref{eqn:reverse_mu_t} and \cref{eqn:reverse_sigma_t}.
Accordingly, the learning objective minimizes the noise prediction error:
\begin{align}
\begin{split}
        \mathcal{L}_\theta & = \mathbb{E}_{t,\boldsymbol{q}^t,\epsilon^t}\left[\|\epsilon^t - \epsilon_\theta(\boldsymbol{q}^t, \mathcal{X}, t)\|^2\right] \\
        & = \mathbb{E}_{t,\boldsymbol{q}^0,\epsilon^t}\left[\left\|\epsilon^t - \epsilon_\theta\left(\sqrt{\bar{\alpha}^t}\boldsymbol{q}^0 + \sqrt{1-\bar{\alpha}^t}\epsilon^t, \mathcal{X}, t\right)\right\|^2\right],
\end{split}
\end{align}
where $\epsilon^t$ is the Gaussian noise injected at step $t$, $\bar{\alpha}^t = \prod_{i=1}^t(1-\beta^i)$ is computed from the schedule $\beta^i$, and $\epsilon_\theta(\cdot)$ denotes the noise predictor.
This objective trains $\epsilon_\theta(\cdot)$ to recover $\epsilon^t$ from the noisy configuration $\boldsymbol{q}^t$ and conditioning $\mathcal{X}$ at an arbitrary timestep $t$.
We provide the full training procedure in~\cref{alg:training_ikdiffuser}.

Using \cref{eqn:forward_diffusion_nice}, we can express $\boldsymbol{q}^0 = \frac{1}{\sqrt{\bar{\alpha}^t}}\left(\boldsymbol{q}^t - \sqrt{1-\bar{\alpha}^t}\epsilon^t\right)$.
With a trained predictor $\epsilon_\theta$, we estimate the original configuration as:
\begin{align}
    \hat{\boldsymbol{q}}^0(\boldsymbol{q}^t, \mathcal{X}, t) = \frac{1}{\sqrt{\bar{\alpha}^t}}\left(\boldsymbol{q}^t - \sqrt{1-\bar{\alpha}^t} \epsilon_\theta(\boldsymbol{q}^t, \mathcal{X}, t)\right),
\end{align}
and define the learned reverse transition by substituting $\hat{\boldsymbol{q}}^0$ into \cref{eqn:reverse_mu_t}.
This yields the parameterization of the reverse denoising transition $p_\theta(\boldsymbol{q}^{t-1} | \boldsymbol{q}^t, \mathcal{X})$ in \cref{eqn:rev_diffusion}:
\begin{align}
\label{eqn:mean_var_diffusion}
\begin{split}
    \boldsymbol{\mu}_\theta(\boldsymbol{q}^t, \mathcal{X}, t) & = \frac{1}{\sqrt{\alpha^t}}\left(\boldsymbol{q}^t - \frac{1-\alpha^t}{\sqrt{1-\bar{\alpha}^t}}\epsilon_\theta\left(\boldsymbol{q}^t, \mathcal{X}, t\right)\right)  \\
    \Sigma_\theta(\boldsymbol{q}^t, \mathcal{X}, t) & = \frac{1-\bar{\alpha}_{t-1}}{1-\bar{\alpha}_t} \beta_t \mathbf{I},
\end{split}
\end{align}
where $\alpha^t = 1-\beta^t$ and $\bar{\alpha}^t = \prod_{i=1}^t \alpha^i$ are determined by the noise schedule $\beta^t$.
Following~\cite{ho2020denoising}, we keep $\Sigma_\theta$ fixed rather than learn it to improve training stability and sampling quality. 
The resulting denoising transition $p_\theta(\boldsymbol{q}^{t-1} | \boldsymbol{q}^t, \mathcal{X})$ is then used to sample \ac{ik} solutions conditioned on the desired end-effector poses $\mathcal{X}$ and task-specific constraints $\mathcal{J}$.

\begin{algorithm}[tb!]
    \caption{Training IKDiffuser}
    \label{alg:training_ikdiffuser}
    \LinesNumbered
    \SetKwInOut{KIN}{Input}
    \SetKwInOut{KOUT}{Output}
    
    \KIN{Joint states and end effector poses$\{(\boldsymbol{q}^0, \mathcal{X})\}$}
    \KOUT{Learned noise prediction model $\epsilon_\theta(\boldsymbol{q}^t, \mathcal{X}, t)$}
    
    \Repeat{converged}{
        \nonl\nosemic \textcolor{blue}{// draw a sample from the data set} \;
        $(\boldsymbol{q}^0, \mathcal{X}) \sim \{(\boldsymbol{q}^0, \mathcal{X}), ...\}$ \;
        \nonl\nosemic \textcolor{blue}{// randomly select a time step for forward process} \;
        $t \sim \text{Uniform}(\{1, \dots, T\})$ \;
        \nonl\nosemic \textcolor{blue}{// sample noise from a standard Gaussian} \;
        $\epsilon \sim \mathcal{N}(\boldsymbol{0}, \mathbf{I})$ \;
        \nonl\nosemic \textcolor{blue}{// corrupt the data over $t$ forward steps} \;
        $\boldsymbol{q}^t = \sqrt{\bar{\alpha}_t} \boldsymbol{q}^0 + \sqrt{1 - \bar{\alpha}_t} \epsilon$ \;
        \nonl\nosemic \textcolor{blue}{// take the gradient step on the noise prediction model} \;
        $\theta = \theta - \eta \nabla_\theta \| \epsilon - \epsilon_\theta(\boldsymbol{q}^t, \mathcal{X}, t) \|^2$
    }
    \nosemic \Return $\epsilon_\theta$ \;
\end{algorithm}

\begin{figure*}[t!]
    \centering
    \includegraphics[width=\linewidth]{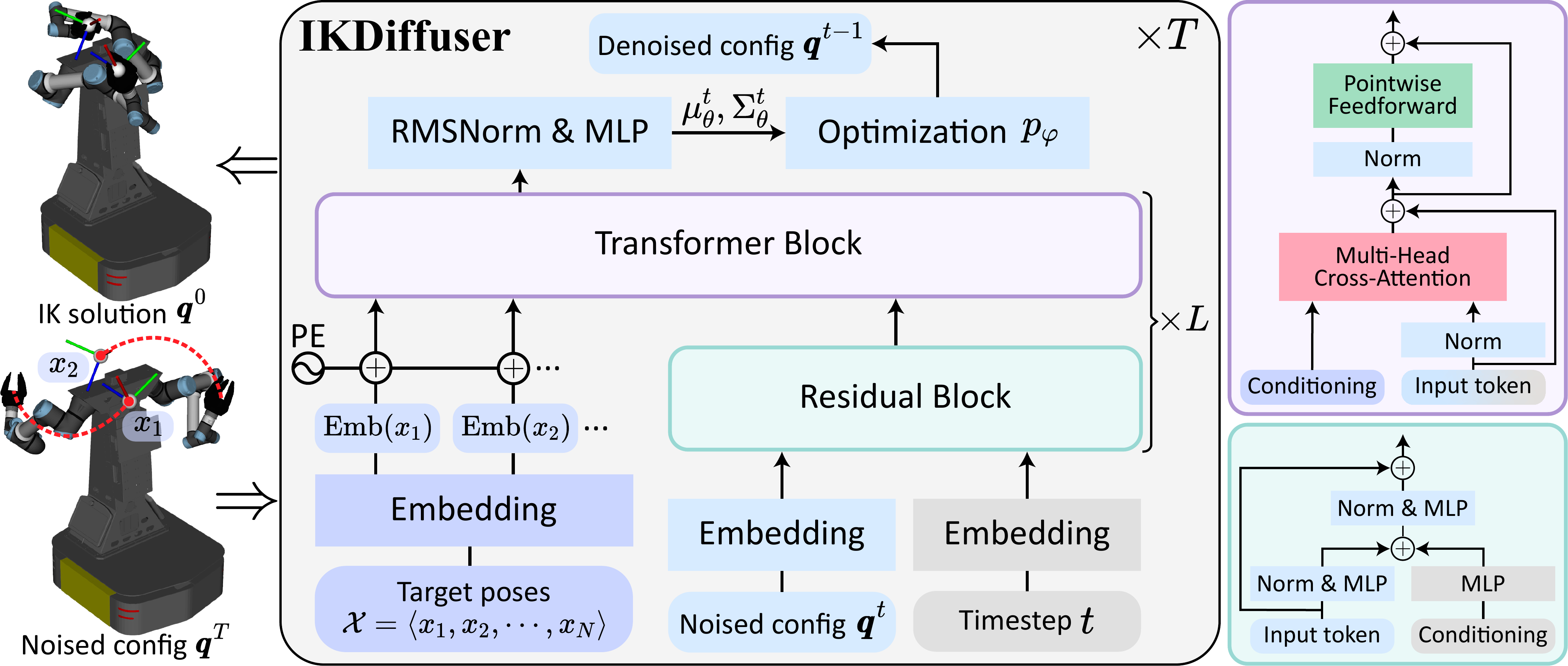}
     \caption{\textbf{Architecture of \textsc{IKDiffuser}.} The model generates inverse kinematics solutions $\boldsymbol{q}^0$ by iteratively denoising Gaussian noise $\boldsymbol{q}^T$ over $T$ timesteps, conditioned on target end-effector poses $\mathcal{X}$. Each end-effector pose $\boldsymbol{x}_i$ is embedded with positional encoding (PE), while timestep $t$ is integrated with denoised configuration $\boldsymbol{q}^t$ through a Residual block. The Transformer block employs cross-attention to learn the relation between joint configurations and end-effector poses, with detailed block structures shown in the colored boxes on the right.}
    \label{fig:architecture}
\end{figure*}

\textbf{Model Architecture:} 
The network architecture of the noise prediction model $\epsilon_\theta(\cdot)$ builds upon conditional diffusion models~\cite{peebles2023scalable, huang2023diffusion}, with a network design that accommodates variable numbers of end effectors, making it adaptable to diverse multi-arm systems. 
Unlike prior work~\cite{ames2022ikflow}, which concatenates end-effector poses into a single conditional input, we treat end-effector poses as a sequence of tokens, where each token corresponds to an individual end-effector pose $\boldsymbol{x}_i$. 
This design enables the model to better capture the relationships between joint configurations and end-effector poses (see \cref{subsec:ablation} for the ablation study).
As illustrated in \cref{fig:architecture}, its core is a \textit{Transformer Block} with multi-head cross-attention~\cite{vaswani2017attention} to model dependencies between joint configurations and end-effector poses. 
Each end-effector pose $\boldsymbol{x}_i$ is encoded with positional encoding (PE) to form the keys and values, while the query is obtained from a time-conditional noised configuration produced by a \textit{Residual Block} that jointly embeds the noised joint configuration $\boldsymbol{q}^t$ and timestep $t$.
Since the joint configuration is a single ``token'' sequence, self-attention is unnecessary in the  \textit{Transformer block}~\cite{vaswani2017attention}. 
The cross-attention mechanism alone captures the correlation between joint configurations and end-effector poses, simplifying the model complexity and improving efficiency.
At each inference step, the model estimates the diffusion noise $\epsilon_\theta^t$ by applying cross-attention between the target end-effector poses and the time-conditional noised configuration.
Given this estimate $\epsilon_\theta^t$, it then computes the mean $\boldsymbol{\mu}_\theta^t$ and variance $\Sigma_\theta^t$ of the reverse diffusion prior according to \cref{eqn:mean_var_diffusion}.
Finally, the denoised joint configuration $\boldsymbol{q}^{t-1}$ is sampled following \cref{eqn:cond_rev_process_full}, where the gradient $\boldsymbol{g}$ steers the process toward satisfying the optimization objectives.

\subsection{Objective-guided Sampling} \label{subsec:guided_sampling}

To solve the \ac{ik} problem, \name samples from the posterior distribution $p(\boldsymbol{q}^0 | \mathcal{X}, \mathcal{J})$, which is decomposed into the diffusion prior $p_{\theta}(\boldsymbol{q}^0 | \mathcal{X})$ and the task-specific constraint likelihood $p_{\varphi}(\mathcal{J} | \boldsymbol{q}^0, \mathcal{X})$. 
We incorporate the differentiable objectives $\mathcal{J}$ into the reverse diffusion process by combining the learned denoising prior $p_\theta(\boldsymbol{q}^{t-1} | \boldsymbol{q}^t, \mathcal{X})$ with the constraint likelihood at each timestep $t$:
\begin{align}
\label{eqn:cond_rev_process}
p(\boldsymbol{q}^{t-1}|\boldsymbol{q}^t, \mathcal{X}, \mathcal{J})\ \propto\ p_\theta(\boldsymbol{q}^{t-1}|\boldsymbol{q}^t, \mathcal{X}) p_\varphi(\mathcal{J} | \boldsymbol{q}^{t-1}, \mathcal{X}).
\end{align}

Given that $p_\theta(\boldsymbol{q}^{t-1} | \boldsymbol{q}^t, \mathcal{X})$ concentrates its mass near the mean $\boldsymbol{\mu}_\theta^t = \boldsymbol{\mu}_\theta(\boldsymbol{q}^t, \mathcal{X}, t)$, we approximate $p_{\varphi}(\mathcal{J} | \boldsymbol{q}^0, \mathcal{X})$ using a first-order Taylor expansion around $\boldsymbol{\mu}_\theta^t$:
\begin{align}
\label{eqn:approx_constraint}
\begin{split}
\log p_{\varphi}(\mathcal{J} | \boldsymbol{q}^{t-1}, \mathcal{X}) \approx \log p_{\varphi}(\mathcal{J} | \boldsymbol{\mu}_\theta^t, \mathcal{X}) + (\boldsymbol{q}^{t-1} - \boldsymbol{\mu}_\theta^t) \boldsymbol{g} \\
\text{where}~\boldsymbol{g} = \nabla_{\boldsymbol{q}^{t-1}} \log p_{\varphi}(\mathcal{J} | \boldsymbol{q}^{t-1}, \mathcal{X})|_{\boldsymbol{q}^{t-1}=\boldsymbol{\mu}_\theta^t},
\end{split}    
\end{align}
where $\log p_{\varphi}(\mathcal{J} | \boldsymbol{\mu}_\theta^t, \mathcal{X})$ is a constant, $\boldsymbol{\mu}_\theta^t = \boldsymbol{\mu}_\theta(\boldsymbol{q}^t, \mathcal{X}, t)$ is the inferred parameters of original diffusion process (see~\cref{eqn:mean_var_diffusion}).
Of note, although the noise schedule $\beta^t$ is larger at early steps during the diffusion process and may increase deviation from the expansion point $\boldsymbol{\mu}_\theta^t$, it remains small in our setup, with a maximum of no more than 0.04 (and often below 0.01). This keeps the Taylor approximation accurate even at early timesteps and across diverse kinematic complexities, consistent with both prior analysis~\cite{carvalho2023motion} and our empirical results. Moreover, since the objective distribution is typically a product of multiple probability terms, applying the logarithm in~\cref{eqn:approx_constraint} converts it into a summation of components, improving tractability and numerical stability by avoiding unstable derivatives of products in high-dimensional, low-probability settings.

\setstretch{1.}

Then, substituting \cref{eqn:approx_constraint} into \cref{eqn:cond_rev_process} yields the guided transition kernel:
\begin{align}
\label {eqn:cond_rev_process_full}
\begin{split}
    &~\log p(\boldsymbol{q}^{t-1}|\boldsymbol{q}^{t}, \mathcal{X}, \mathcal{J}) \\
    = &~\log p_\theta(\boldsymbol{q}^{t-1}|\boldsymbol{q}^t, \mathcal{X}) + \log p_\varphi(\mathcal{J} | \boldsymbol{q}^{t-1}, \mathcal{X}) \\
    \propto &~-\frac{1}{2}(\boldsymbol{q}^{t-1} - \boldsymbol{\mu}_\theta^{t})^{\top}{\Sigma^{t}_\theta}^{-1}(\boldsymbol{q}^{t-1} - \boldsymbol{\mu}_\theta^{t}) + (\boldsymbol{q}^{t-1} - \boldsymbol{\mu}_\theta^{t})\boldsymbol{g} \\
    \propto &~-\frac{1}{2}(\boldsymbol{q}^{t-1} - \boldsymbol{\mu}_\theta^{t} - \Sigma_\theta^{t}\boldsymbol{g})^{\top}{\Sigma^{t}_\theta}^{-1}(\boldsymbol{q}^{t-1} - \boldsymbol{\mu}_\theta^{t} - \Sigma_\theta^{t}\boldsymbol{g}) \\
    = &~\log \mathcal{N}(\boldsymbol{q}^{t-1}; \boldsymbol{\mu}_\theta^t + \Sigma_\theta^{t}\boldsymbol{g}, \Sigma_\theta^{t}).
\end{split}    
\end{align}
This derivation shows that sampling from $p(\boldsymbol{q}^{t-1}|\boldsymbol{q}^{t}, \mathcal{X}, \mathcal{J})$ is equivalent to sampling from a Gaussian distribution $ \mathcal{N}(\boldsymbol{q}^{t-1}; \boldsymbol{\mu}_\theta^t + \Sigma_\theta^{t}\boldsymbol{g}, \Sigma_\theta^{t})$. 
Here, the gradient $\boldsymbol{g}$ acts as a guidance signal, steering the denoising trajectory toward the objective $\mathcal{J}$. 
The complete guided sampling procedure is summarized in \cref{alg:sampling_ikdiffuser}.

\begin{algorithm}[t!]
    \caption{Sampling via IKDiffuser}
    \label{alg:sampling_ikdiffuser}
    \LinesNumbered
    \SetKwInOut{KIN}{Input}
    \SetKwInOut{KOUT}{Output}
    
    \KIN{End-effector poses $\mathcal{X}$}
    \KOUT{Joint configuration $\boldsymbol{q}^0$}
    
    $\boldsymbol{q}^T \sim \mathcal{N}(\boldsymbol{0}, \mathbf{I})$

    \For{$t = T, T-1, \cdots, 1$}{
        \nonl\nosemic \textcolor{blue}{// compute the diffusion prior in terms of \cref{eqn:mean_var_diffusion}} \;
        $\boldsymbol{\mu}^t = \boldsymbol{\mu}_\theta(\boldsymbol{q}^t, \mathcal{X}, t), \Sigma^t = \Sigma_\theta(\boldsymbol{q}^t, \mathcal{X}, t)$ \;
        \nonl\nosemic \textcolor{blue}{// compute the gradient of the objective constraints} \;
        $\boldsymbol{g}^t = \nabla_{\boldsymbol{q}} \log p_{\varphi}(\mathcal{J} | \boldsymbol{q}, \mathcal{X})|_{\boldsymbol{q}=\boldsymbol{\mu^t}}$ \;
        \nonl\nosemic \textcolor{blue}{// denoising at step $t$} \;
        $\boldsymbol{q}^{t-1} \sim \mathcal{N}(\boldsymbol{\mu}^t + \Sigma^t \boldsymbol{g}^t, \Sigma^t)$ \;
    }
    \nosemic \Return $\boldsymbol{q}^0$ \;
\end{algorithm}
\vspace{-2pt}



\subsection{Task-specific Objectives}
\label{subsec:objective}

In many practical settings, \ac{ik} solutions are not uniquely characterized by end-effector pose constraints alone: different tasks may favor different regions of the configuration space, for example, prioritizing continuity with previous solutions, maximizing manipulability, or satisfying additional safety or comfort constraints.
Instead of retraining a separate model for every preference, we equip \name with a set of task-specific objectives that can be injected at sampling time as auxiliary likelihood terms whose gradients guide the diffusion process.
In general, \emph{users can incorporate additional, customized objectives as long as they are differentiable} with respect to $\boldsymbol{q}$, enabling plug-and-play control of the generated \ac{ik} solutions without retraining the model. 
Specifically, we define the overall task-specific objective likelihood as the product of multiple independent objectives:
\begin{align}
    p_{\varphi}(\mathcal{J} | \boldsymbol{q}^{t-1}, \mathcal{X}) = \prod_{k} p_{\varphi}(\mathcal{J}_k | \boldsymbol{q}^{t-1}, \mathcal{X}),
\end{align}
where each objective $\mathcal{J}_k$ contributes a separate likelihood term. 
Accordingly, in terms of \cref{eqn:approx_constraint}, the overall gradient used in \cref{alg:sampling_ikdiffuser} is the sum of the individual objective gradients:
\begin{align}
    \boldsymbol{g} = \sum_k \nabla_{\boldsymbol{q}^{t-1}} \log p_{\varphi}(\mathcal{J}_k | \boldsymbol{q}^{t-1}, \mathcal{X})|_{\boldsymbol{q}=\boldsymbol{\mu^t}}.
\end{align}

In this article, we present warm-start and manipulability as two representative objectives to illustrate how task-specific preferences can be adopted within this formulation.
These examples highlight the general mechanism by which \name steers the generation of \ac{ik} solutions toward joint configurations that satisfy task-specified requirements, while still meeting the original \ac{ik} constraints, via the objective-guided sampling introduced in \cref{subsec:guided_sampling}.
As such, this framework could naturally extend to additional constraints and customized objectives, enabling flexible adaptation to diverse tasks without modifying or retraining the base model.

\textbf{Warm-started IK Generation:} 
In \ac{ik}-based applications such as motion planning, trajectory following, and teleoperation, using prior solutions as seeds is crucial for ensuring smooth, stable motion and consistent behavior.
This strategy reduces abrupt joint angle changes and, in redundant systems like dual-arm robots where the \ac{ik} problem admits infinitely many solutions, steers the solver toward specific, repeatable configurations that are preferred in practice.
To encourage \name to generate \ac{ik} solutions that preserve this continuity, we introduce a warm-start objective that penalizes deviation from a prior solution:
\begin{align}
    \log p_{\varphi}(\mathcal{J}_{\text{warm}}|\boldsymbol{q}, \mathcal{X}) = -\|\boldsymbol{q} - \boldsymbol{q}^{\text{prior}}\|^2_2
\end{align}
where $\boldsymbol{q}^{\text{prior}}$ represents the prior solution given in the warm-start objective $\mathcal{J}_{\text{warm}}$.
The likelihood increases as the generated joint configuration $\boldsymbol{q}$ approaches the prior solution $\boldsymbol{q}^{\text{prior}}$.
The corresponding objective gradient at time step $t$ is:
\begin{align}
    \boldsymbol{g}^t_{\text{warm}} = \nabla_{\boldsymbol{q}} \|\boldsymbol{q} - \boldsymbol{q}^{\text{prior}}\|^2|_{\boldsymbol{q}=\mu^t} = -2(\mu^t - \boldsymbol{q}^{\text{prior}})
\end{align}
Incorporating this gradient into the sampling process (see~\cref{alg:sampling_ikdiffuser}) yields solutions that remain close to prior configurations, enabling smooth and stable motions in practical applications without retraining the model.

\textbf{Manipulability-aware IK Generation:} 
In robotic manipulation, optimizing manipulability is essential for avoiding singularities and ensuring robust execution. 
High-manipulability configurations enhance a robot's ability to exert forces and velocities in task-relevant directions, particularly in redundant systems, thereby reducing the risk of singularities, improving adaptability to unexpected disturbances, and enabling smoother trajectory execution.
However, traditional \ac{ik} approaches often prioritize positional accuracy without explicitly optimizing for manipulability, which can yield mathematically valid but kinematically constrained solutions. 
To encourage \name to generate high-manipulability solutions, we introduce a manipulability-aware objective that maximizes the manipulability of the generated joint configurations:
\begin{align}
    \log p_{\varphi}(\mathcal{J}_{\text{manip}}|\boldsymbol{q}, \mathcal{X}) = \sum_i^{N_\text{ee}} \sqrt{\det(\boldsymbol{J}_i\boldsymbol{J}_i^T)}
    \label{eqn:manipulability}
\end{align}  
where the right-hand side is the sum of manipulability measures across all serial chains of each end-effector. 
Specifically, $N_\text{ee}$ is the number of end effectors, and $\boldsymbol{J}_i$ is the Jacobian matrix of the $i$-th end-effector at configuration $\boldsymbol{q}$. 
Following the formulation in~\cite{wang2016whole}, we compute the gradient of \cref{eqn:manipulability} with respect to $\boldsymbol{q}$ and use it to guide sampling in \cref{alg:sampling_ikdiffuser}.


\subsection{Marginal Inference}
\label{subsec:marginal}

While \name is trained to model the conditional distribution $p_\theta(\boldsymbol{q}^0 \mid \mathcal{X})$ over joint configurations given all end-effector poses $\mathcal{X} = \langle \boldsymbol{x}_1, \boldsymbol{x}_2, \dots,\boldsymbol{x}_N \rangle$, many practical tasks only constrain a subset of end-effectors. For example, a humanoid manipulation task may only require arm poses to be specified, leaving the feet and head free. In such cases, the desired posterior takes the form $p(\boldsymbol{q}^0 \mid \mathcal{X}_S, \mathcal{J})$, where $S \subseteq \{1,\dots,N\}$ denotes the indices of constrained end-effectors. Formally, this corresponds to marginalizing out the unconstrained poses:
\begin{align}
    p(\boldsymbol{q}^0 \mid \mathcal{X}_S, \mathcal{J})
    \propto
    \int p(\boldsymbol{q}^0 \mid \mathcal{X}, \mathcal{J}) \, p(\mathcal{X}_{\bar{S}} \mid \mathcal{X}_S, \mathcal{J}) \, d\mathcal{X}_{\bar{S}},
\end{align}
where $\bar{S}$ is the complement of $S$. Evaluating this integral is intractable for high-dimensional space, as it requires integrating over all possible poses for the unconstrained end-effectors.

Instead of explicit integration, we induce the marginalization at the model level by modifying the conditioning distribution during training. Let $\mathcal{X} = \langle \boldsymbol{x}_1, \boldsymbol{x}_2, \dots,\boldsymbol{x}_N \rangle$ denote the ordered pose sequence associated with a configuration $\boldsymbol{q}^0$. We augment the conditioning space with a dedicated ``empty'' symbol $\boldsymbol{x}_{\emptyset}$ that represents the absence of a constraint on a particular end-effector. During training, we apply \emph{conditioning dropout}: for each pair $(\boldsymbol{q}^0, \mathcal{X})$, we sample a binary mask $m \in \{0,1\}^N$ and construct a masked conditioning sequence $\tilde{\mathcal{X}}$ according to
\begin{align}
    \tilde{\boldsymbol{x}}_i =
    \begin{cases}
        \boldsymbol{x}_i, & m_i = 1, \\
        \boldsymbol{x}_{\emptyset}, & m_i = 0,
    \end{cases}
\end{align}
where each entry is independently dropped with probability $p_{\text{drop}} \in [0,1)$. 
The diffusion network is then trained with the standard noise-prediction objective, but always conditioned on $\tilde{\mathcal{X}}$ rather than the fully specified $\mathcal{X}$, thereby learning a family of conditional distributions
\begin{align}
    p_\theta(\boldsymbol{q}^0 \mid \tilde{\mathcal{X}}, \mathcal{J}), \quad
    \tilde{\mathcal{X}} \in \{\boldsymbol{x}_1, \boldsymbol{x}_ 2, \dots,\boldsymbol{x}_N,\boldsymbol{x}_{\emptyset}\}^N,
\end{align}
where masked entries correspond to unconstrained end-effectors. However, directly encoding the empty symbol $\boldsymbol{x}_{\emptyset}$ as a zero vector can be problematic, as it is mathematically indistinguishable from a valid target pose located at the origin. 
To resolve this ambiguity, we parameterize $\boldsymbol{x}_{\emptyset}$ as a learnable latent code $z_\emptyset$ within the embedding space of the pose encoder.
The $z_{\emptyset}$ is initialized randomly and optimized jointly with all other network parameters via backpropagation. 
Because $z_{\emptyset}$ participates in the loss whenever conditioning dropout masks out one or more end-effectors, the model learns an embedding consistently associated with ``no constraint'' rather than any particular geometric pose, which empirically yields more stable behavior under partial conditioning.

At inference time, a marginal IK query specified by $\mathcal{X}_S$ is instantiated by constructing a mixed conditioning sequence $\tilde{\mathcal{X}} = [\tilde{\boldsymbol{x}}_1,\tilde{\boldsymbol{x}}_2, \dots,\tilde{\boldsymbol{x}}_N]$ with
\begin{align}
    \tilde{x}_i =
    \begin{cases}
        \boldsymbol{x}_i, & i \in S, \\
        z_{\emptyset}, & i \notin S.
    \end{cases}
    \label{eqn:mixed_conditioning}
\end{align}
Feeding this partially specified sequence into the reverse diffusion process yields samples from $p_\theta(\boldsymbol{q}^0 \mid \tilde{\mathcal{X}}, \mathcal{J})$, which approximate the desired $p(\boldsymbol{q}^0 \mid \mathcal{X}_S, \mathcal{J})$: end-effectors in $S$ are driven toward their prescribed targets, while the remaining limbs are placed in configurations that are statistically consistent with the training distribution. Thus, conditioning dropout together with a learned latent empty token provides a unified mechanism for handling fully constrained, partially constrained, and unconstrained IK queries within a single \name{} model, effectively implementing marginal inference over missing end-effector poses. We demonstrate this capability in our experiments (see \cref{sec:experiment:marginal}).

\section{Experiment}\label{sec:experiment}

We evaluate \name across eight robotic configurations (comprising seven distinct robot models shown in \cref{fig:teaser} and one variant of Baxter) with diverse kinematic structures. These systems range from fixed-base manipulators and dual-arm mobile manipulators to dexterous hands and humanoids. Our experiments demonstrate the advantages of \name over both optimization- and learning-based baselines.

Our study investigates six key aspects of \name:
\begin{itemize}[leftmargin=*]
    \item The diversity and accuracy of generated solutions compared to state-of-the-art learning-based IK solvers (\cref{sec:experiment:ik_gen});
    \item The efficiency and effectiveness of using generated solutions to seed optimization-based IK solvers (\cref{sec:experiment:seed});
    \item The computational efficiency and scalability across batch sizes, comparing native optimization solvers against those seeded by the proposed model (\cref{sec:experiment:time});
    \item The adaptability to task-specific objectives via the incorporation of additional cost functions at inference time, without retraining (\cref{sec:experiment:objective});
    \item The capability to handle partially specified end-effector goals via masked marginal inference, without retraining the model (\cref{sec:experiment:marginal});
    \item An error analysis investigating how specific kinematic parameters influence model performance (\cref{sec:experiment:error}).
\end{itemize}


\subsection{Experimental Setup}\label{sec:experiment:setup}
\textbf{Data Generation:} 
To construct the dataset, we uniformly sampled joint configurations within joint limits for each robotic system. Forward kinematics was then applied to compute the corresponding Cartesian poses of all end-effectors. Configurations resulting in self-collisions were discarded to ensure physical feasibility. The dataset consists of end-effector poses as conditional inputs paired with joint configurations as target outputs. 
For training \name and all baselines, we generated 10M samples for each robotic platform.

To accelerate the data generation process, we developed a GPU-accelerated toolchain based on the cuRobo library~\cite{sundaralingam2023curobo}, enabling efficient parallel forward kinematics and collision checking over large batches.
Given a robot URDF as input, the toolchain outputs valid joint configurations and their corresponding end-effector poses, with the number of samples specified by the user.
Using an NVIDIA RTX 4090 GPU, this toolchain could generate 10M samples within 30 seconds.

\textbf{Implementation and Hyperparameters:} 
\name adopts a Transformer-based architecture (27.2M parameters), comprising eight stacked residual–Transformer blocks with eight attention heads, a hidden dimension of 512, and a feed-forward dimension of 1024 (see~\cref{fig:architecture}). Encoders employ GeLU activation~\cite{hendrycks2016gaussian} with RMSNorm~\cite{zhang2019root}, and all Transformer blocks use a pre-norm design~\cite{xiong2020layer} for stable training. For each robotic system, training was performed on an NVIDIA RTX 4090 GPU for 200 epochs, using the AdamW optimizer and a constant learning rate of $10^{-4}$. A linear noise schedule was applied, increasing from $10^{-4}$ to $0.04$ across 100 diffusion steps. During inference, we used DPM-Solver++~\cite{lu2022dpm} to reduce sampling from 100 steps to 5. 

\textbf{Baselines:}
We benchmarked \name against three state-of-the-art approaches: IKFlow~\cite{ames2022ikflow}, a normalizing-flow-based IK solver; Pink~\cite{pink2025}, an optimization-based solver that leverages Pinocchio~\cite{carpentier2019pinocchio} and quadratic programming; and cuRobo~\cite{sundaralingam2023curobo}, a GPU-accelerated solver that applies the evolutionary strategy to IK for kinematic trees (except for serial chains). These baselines were selected to represent the three dominant paradigms: learning-based (generative), gradient-based, and gradient-free (GPU-paralleled) methods, thereby ensuring a comprehensive comparison. 
To extend IKFlow for our problem, we concatenated end-effector poses as conditional inputs and significantly tuned the architecture to maximize performance on kinematic trees. Specifically, we adopted a scaled-up design with 16 coupling blocks and 1024 hidden units per layer. This resulted in a 67.8M-parameter model, explicitly optimized to capture the high-dimensional complexity of multi-arm kinematics.

\begin{table*}[th!]
    \centering
    \caption{\textbf{Task 1: }Comparison of IKFlow and \name.}
    \resizebox{\textwidth}{!}{%
    \setlength{\tabcolsep}{4.5pt} 
    \setstretch{1.2} 
    \begin{tabular}{lcc >{\columncolor{red!20}}c >{\columncolor{red!20}}c >{\columncolor{red!20}}c >{\columncolor{red!20}}c >{\columncolor{blue!20}}c >{\columncolor{blue!20}}c >{\columncolor{blue!20}}c >{\columncolor{blue!20}}c}
    \toprule
    \multicolumn{11}{c}{\textbf{Task 1: IK Solution Generation}} \\
    \toprule
    \multicolumn{3}{c}{\textbf{Robot Specifications}} & \multicolumn{4}{>{\columncolor{red!20}}c}{\textbf{IKFlow [67.8M]}} & \multicolumn{4}{>{\columncolor{blue!20}}c}{\textbf{\name [27.2M]}} \\
    \cmidrule(lr){1-3} \cmidrule(lr){4-7} \cmidrule(lr){8-11}
    Robot Platform & \textbf{$N_\text{ee}$} & DoF & Coll. (\%)$\downarrow$ & Position (mm)$\downarrow$ & Angular (deg)$\downarrow$ & Div.$\uparrow$ & Coll. (\%)$\downarrow$ & Position (mm)$\downarrow$ & Angular (deg)$\downarrow$ & Div.$\uparrow$ \\
    \midrule
    Franka Research 3 & 1 & 7 & 1.51 & $15.12 \pm 36.17$ & $3.78 \pm 11.82$ & 3.20 & \textbf{0.49} & $\textbf{4.90} \pm \textbf{2.84}$ & $\textbf{1.05} \pm \textbf{0.62}$ & \textbf{3.33} \\
    Baxter Dualarm & 2 & 14 & 7.61 & $57.48 \pm 63.45$ & $14.65 \pm 16.89$ & 1.42 & \textbf{1.78} & $\textbf{5.25} \pm \textbf{2.09}$ & $\textbf{1.22} \pm \textbf{0.51}$ & \textbf{1.50} \\
    Mobile Dual UR5e & 2 & 15 & 44.34 & $363.82 \pm 148.09$ & $119.02 \pm 28.42$ & 1.17 & \textbf{11.18} & $\textbf{9.45} \pm \textbf{4.16}$ & $\textbf{1.48} \pm \textbf{0.68}$ & \textbf{1.28} \\
    Dual RM76 and Waist & 2 & 17 & 4.21 & $82.37 \pm 60.92$ & $29.90 \pm 23.89$ & 1.63 & \textbf{0.66} & $\textbf{4.46} \pm \textbf{1.74}$ & $\textbf{1.59} \pm \textbf{0.83}$ & \textbf{1.67} \\
    Mobile Baxter Dualarm & 2 & 17 & 13.20 & $89.62 \pm 135.60$ & $16.97 \pm 20.13$ & 1.64 & \textbf{3.72} & $\textbf{9.44} \pm \textbf{3.66}$ & $\textbf{1.64} \pm \textbf{0.73}$ & \textbf{1.69} \\
    LEAP Hand Float Base & 4 & 22 & 31.46 & $19.10 \pm 35.68$ & $10.22 \pm 20.86$ & 1.23 & \textbf{2.85} & $\textbf{1.73} \pm \textbf{0.72}$ & $\textbf{0.93} \pm \textbf{0.33}$ & \textbf{1.39} \\
    NASA Robonaut & 4 & 29 & 9.51 & $73.06 \pm 98.30$ & $13.27 \pm 14.85$ & 1.28 & \textbf{3.20} & $\textbf{9.28} \pm \textbf{3.65}$ & $\textbf{1.51} \pm \textbf{0.57}$ & \textbf{1.42} \\
    Unitree G1 & 4 & 29 & 15.36 & $20.01 \pm 33.41$ & $7.24 \pm 10.69$ & 1.52 & \textbf{6.35} & $\textbf{5.31} \pm \textbf{1.56}$ & $\textbf{1.48} \pm \textbf{0.49}$ & \textbf{1.65} \\
    \bottomrule
    \end{tabular}%
    }%
    \label{tab:ik_quality}
\end{table*}

\setstretch{1.03}

\subsection{Task 1: IK Solution Generation} \label{sec:experiment:ik_gen}

\textbf{Setup:} 
For each robotic platform being evaluated, we generate a test set of previously unseen 10,000 end-effector poses by uniformly sampling the joint space. Each model produces 100 candidate IK solutions per pose for diversity assessment. Forward kinematics is then used to compute actual end-effector poses, which are compared against target end-effector poses to measure solution quality. All baselines are trained on the same datasets as \name to ensure fairness.

\textbf{Metrics:}  
Four evaluation metrics are adopted: (i) \emph{self-collision rate}, the percentage of generated solutions resulting in collisions; (ii) \emph{position error}, measured as the Euclidean distance between generated and target positions; (iii) \emph{angular error}, defined as the geodesic distance between normalized quaternions; (iv) \emph{diversity score}, computed as the ratio of the average pairwise Euclidean distance of solutions generated by \name to those produced by an optimization-based solver (cuRobo); a score greater than 1.0 indicates higher diversity compared to the baseline.

\textbf{Results and Analysis:}
\cref{tab:ik_quality} shows the results. 
Across all systems, \name consistently yields lower collision rates, often reducing them by 3 to 10 times relative to IKFlow (\eg, 1.78\% vs.\ 7.61\% on Baxter Dualarm; 2.85\% vs.\ 31.46\% on LEAP Hand). This improvement demonstrates its ability to encode feasibility implicitly without adding explicit collision constraints during inference.
In addition, \name achieves substantially lower position and angular errors, often reducing error magnitudes by one order, while also exhibiting smaller variances, indicating greater robustness and reliability across sampled solutions. 

In terms of solution diversity, both methods maintain diversity scores above $1.0$, validating their ability to produce more diverse configurations than optimization-based methods with random seeds. 
Notably, \name surpasses IKFlow in most cases, while preserving accuracy and collision avoidance. 
In contrast, IKFlow's diversity often comes at the expense of solution quality, as evidenced by its high errors in position and angular metrics. This demonstrates that \name can balance solution diversity with precision.
Overall, these results highlight three key findings: (i) \name scales effectively to complex, high-DoF kinematic trees where IKFlow collapses, (ii) it consistently reduces collision rates and errors across all systems while maintaining solution diversity, and (iii) its stability across robots ensures predictable and robust performance, making it suitable for practical applications.  

\begin{table*}[th!]
    \centering
    \caption{\textbf{Task 2:} Comparison of seeding cuRobo with IKFlow and \name.}
    \resizebox{\textwidth}{!}{%
    \setlength{\tabcolsep}{4.5pt} 
    \setstretch{1.2} 
    \begin{tabular}{lcc|>{\columncolor{green!10}}c>{\columncolor{green!10}}c>{\columncolor{green!10}}c|>{\columncolor{red!20}}c>{\columncolor{red!20}}c>{\columncolor{red!20}}c|>{\columncolor{blue!20}}c>{\columncolor{blue!20}}c>{\columncolor{blue!20}}c}
        \toprule
        \multicolumn{12}{c}{\textbf{Task 2: Seeding Optimization-based IK Solver (cuRobo)}} \\
        \toprule
        \multicolumn{3}{c|}{\textbf{Robot Specifications}} & \multicolumn{3}{c|}{\cellcolor{green!20}\textbf{cuRobo [200 iter]}} & \multicolumn{3}{c|}{\cellcolor{red!20}\textbf{IKFlow + cuRobo [200 iter]}} & \multicolumn{3}{c}{\cellcolor{blue!20}\textbf{\name + cuRobo [200 iter]}} \\
        \cmidrule(lr){1-3} \cmidrule(lr){4-6} \cmidrule(lr){7-9} \cmidrule(lr){10-12}
        Robot Platform & $N_\text{ee}$ & DoF & Suc. (\%)$\uparrow$ & Pos. (mm)$\downarrow$ & Ang. (deg)$\downarrow$ & Suc. (\%)$\uparrow$ & Pos. (mm)$\downarrow$ & Ang. (deg)$\downarrow$ & Suc. (\%)$\uparrow$ & Pos. (mm)$\downarrow$ & Ang. (deg)$\downarrow$ \\
        \midrule
        Franka Research 3     & 1 & 7  & 74.13 & \textbf{0.0047} & 0.0008 & 94.73 & \textbf{0.0047} & 0.0008 & \textbf{98.47} & \textbf{0.0047} & \textbf{0.0007} \\
        Baxter Dualarm        & 2 & 14 & 44.04 & 0.0081 & 0.0020 & 78.39 & 0.0058 & 0.0013 & \textbf{96.93} & \textbf{0.0053} & \textbf{0.0011} \\
        Mobile Dual UR5e      & 2 & 15 & 35.56 & 0.0139 & 0.0073 & 57.93 & 0.0106 & 0.0056 & \textbf{95.60} & \textbf{0.0043} & \textbf{0.0010} \\
        Dual RM76 and Waist   & 2 & 17 & 83.39 & 0.0052 & 0.0009 & 95.09 & 0.0046 & 0.0008 & \textbf{99.85} & \textbf{0.0043} & \textbf{0.0007} \\
        Mobile Baxter Dualarm & 2 & 17 & 47.52 & 0.0072 & 0.0025 & 91.93 & 0.0046 & 0.0014 & \textbf{97.44} & \textbf{0.0039} & \textbf{0.0007} \\
        LEAP Hand Float Base  & 4 & 22 &  1.46 & 0.0621 & 0.0021 & 17.98 & 0.0508 & 0.0015 & \textbf{80.95} & \textbf{0.0355} & \textbf{0.0011} \\
        NASA Robonaut         & 4 & 29 &  3.30 & 0.0192 & 0.0082 & 64.69 & 0.0093 & 0.0039 & \textbf{93.31} & \textbf{0.0064} & \textbf{0.0024} \\
        Unitree G1            & 4 & 29 & 21.01 & 0.0093 & 0.0019 & 83.06 & 0.0057 & 0.0013 & \textbf{96.96} & \textbf{0.0050} & \textbf{0.0011} \\
        \bottomrule
    \end{tabular}%
    }
    \label{tab:ik_seed}
\end{table*}

\begin{table*}[th!]
    \centering
    \small
    \caption{\textbf{Task 2:} Comparison of seeding Pink with IKFlow and \name.}
    \resizebox{\textwidth}{!}{%
    \setlength{\tabcolsep}{12pt} 
    \setstretch{1.2} 
    \begin{tabular}{lcc|>{\columncolor{green!10}}c>{\columncolor{green!10}}c|>{\columncolor{red!20}}c>{\columncolor{red!20}}c|>{\columncolor{blue!20}}c>{\columncolor{blue!20}}c}
        \toprule
        \multicolumn{9}{c}{\textbf{Task 2: Seeding Optimization-based IK Solver (Pink)}} \\
        \toprule
        \multicolumn{3}{c|}{\textbf{Robot Specifications}} & \multicolumn{2}{c|}{\cellcolor{green!20}\textbf{Pink [200 iter]}} & \multicolumn{2}{c|}{\cellcolor{red!20}\textbf{IKFlow + Pink [200 iter]}} & \multicolumn{2}{c}{\cellcolor{blue!20}\textbf{\name + Pink [200 iter]}} \\
        \cmidrule(lr){1-3} \cmidrule(lr){4-5} \cmidrule(lr){6-7} \cmidrule(lr){8-9}
        Robot Platform & $N_\text{ee}$ & DoF & Suc. (\%)$\uparrow$ & Iter. Count$\downarrow$ & Suc. (\%)$\uparrow$ & Iter. Count$\downarrow$ & Suc. (\%)$\uparrow$ & Iter. Count$\downarrow$ \\
        \midrule
        Franka Research 3     & 1 & 7  & 55.36 & 91.48  & 93.74 & 12.83 & \textbf{98.51} & \textbf{3.01} \\
        Baxter Dualarm        & 2 & 14 & 35.54 & 136.86 & 62.10 & 23.62 & \textbf{95.82} & \textbf{3.14} \\
        Mobile Dual UR5e      & 2 & 15 & 18.09 & 121.04 & 36.80 & 45.33 & \textbf{87.78} & \textbf{1.91} \\
        Dual RM76 and Waist   & 2 & 17 & 69.38 & 78.31  & 94.26 & 8.82  & \textbf{99.04} & \textbf{1.21} \\
        Mobile Baxter Dualarm & 2 & 17 & 40.07 & 98.82  & 64.25 & 5.35  & \textbf{94.86} & \textbf{1.19} \\
        LEAP Hand Float Base  & 4 & 22 & 12.25 & 188.27 & 79.99 & 60.45 & \textbf{96.29} & \textbf{18.84} \\
        NASA Robonaut         & 4 & 29 &  4.15 & 195.31 & 45.14 & 29.55 & \textbf{90.13} & \textbf{6.35} \\
        Unitree G1            & 4 & 29 &  7.65 & 175.87 & 71.77 & 24.74 & \textbf{91.27} & \textbf{6.13} \\
        \bottomrule
    \end{tabular}%
    }
    \label{tab:pink_seed}
\end{table*}

\begin{figure*}[th!]
    \centering
    \includegraphics[width=1\linewidth]{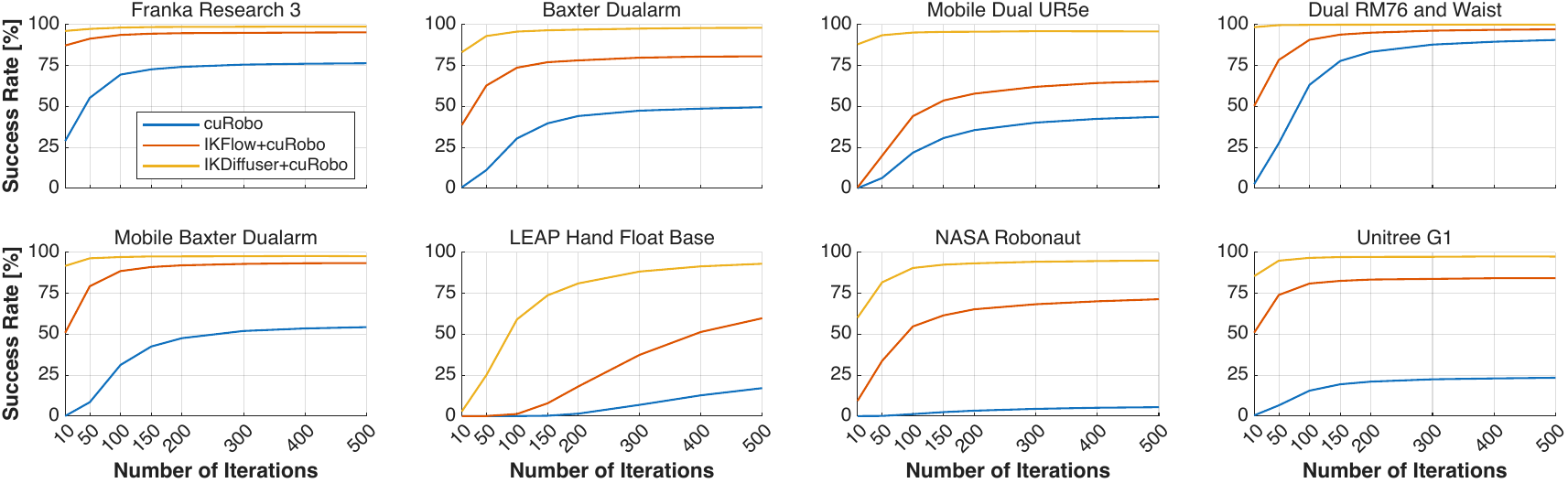}
    \caption{\textbf{Task 2:} Success rate versus optimization iterations, when seeding Optimization-based IK Solver (cuRobo) with \name.}
    \label{fig:time_per_solution}
\end{figure*}

\subsection{Task 2: Seeding Optimization-based IK Solver}\label{sec:experiment:seed}

\textbf{Setup:}  
We assess the effectiveness of seeding two optimization-based solvers, cuRobo and Pink, using the same test set as in Task~1 (10,000 unseen target end-effector poses with 100 runs per pose). Candidate solutions are provided by either IKFlow or \name. Since the solvers adopt different stopping rules---cuRobo runs until the maximum iteration limit is reached, while Pink terminates once the precision threshold is met or exceeded---we employ distinct evaluation metrics tailored to each solver.
Finally, we tune the hyperparameters of both cuRobo and Pink to attain their best achievable performance under our evaluation protocol.

\textbf{Metrics:}  
For \emph{cuRobo}, evaluation includes: (i) \emph{success rate}, the percentage of solutions achieving position error below $0.1$~mm and angular error below $0.01$~radian ($\approx 0.57^\circ$) within 200 iterations; (ii) \emph{position error}, measured as the Euclidean distance between actual and target positions; and (iii) \emph{angular error}, computed via the geodesic distance between normalized quaternions. Both errors are reported at the final iteration.  
For \emph{Pink}, metrics consist of: (i) \emph{success rate}, defined using the same thresholds and iteration cap; and (ii) \emph{iteration count}, the average number of iterations required to reach convergence.  
In addition, for cuRobo, we analyze how the success rate evolves as a function of the maximum allowed iterations.

\textbf{Results and Analysis:}  
\cref{tab:ik_seed,tab:pink_seed} and \cref{fig:time_per_solution} summarize the results. First of all, seeding substantially improves the performance of both optimization-based solvers compared to their vanilla versions. However, the degree of improvement varies significantly depending on the quality of the seeds. IKFlow provides moderate benefits in relatively simple serial systems such as Franka Research 3, where the success rate increases from $74.13\%$ to $94.73\%$. Yet, for more complex multi-arm or humanoid systems, IKFlow's improvements are limited or inconsistent. For instance, in the LEAP Hand with 22 DoFs, the success rate remains below $20\%$ even after seeding, indicating that IKFlow seeds are not accurate enough to guide optimization effectively.

In contrast, \name consistently improves solver performance across all platforms, demonstrating both robustness and scalability. When combined with cuRobo, \name raises success rates above $95\%$ in nearly every case, while simultaneously reducing position and angular errors. These increments are particularly significant for high-DoF and multi-arm systems, where traditional solvers often fail. 
For example, on NASA Robonaut, the success rate increases from $3.30\%$ (vanilla cuRobo) and $64.69\%$ (IKFlow+cuRobo) to $93.31\%$ with \name; on LEAP Hand, \name achieves $80.95\%$ versus $1.46\%$ for vanilla cuRobo.
The convergence plots in \cref{fig:time_per_solution} further reveal that \name enables much faster saturation of success rates, typically within $50\sim100$ iterations, while vanilla cuRobo requires close to the full iteration budget to plateau at significantly lower rates.  

The benefits of \name extend equally to Pink. While vanilla Pink exhibits modest success rates and often requires over $100$ iterations to converge, seeding with IKFlow reduces iteration counts but does not achieve substantial efficacy in multi-arm or humanoid cases. By contrast, \name not only boosts success rates to above $90\%$ in nearly all systems but also reduces the iteration count to less than 10 for most platforms. For instance, in the Baxter Dualarm, iteration counts drop from $136.86$ (vanilla Pink) and $23.62$ (IKFlow + Pink) to only $3.14$ with \name, while success rates climb to $95.82\%$. Similar trends are observed in NASA Robonaut, where \name achieves $90.13\%$ success with an average of just $6.35$ iterations, compared to $4.15\%$ success and nearly $200$ iterations for vanilla Pink. These findings also align with the underlying optimization strategies: gradient-free cuRobo performs better than gradient-based Pink in the absence of good seeds, but once provided with high-quality initialization, Pink converges much faster and avoids poor local minima.  

Overall, these results demonstrate that \name provides high-quality, diverse seeds that significantly improve both the robustness and efficiency of optimization-based IK solvers. The improvements are especially critical in high-DoF, multi-branch systems, where traditional solvers typically fail or require prohibitively many iterations. By accelerating convergence and achieving consistently high success rates, \name enables optimization-based IK methods to become viable for real-time and large-scale robotic applications.  

\setstretch{1.}

\subsection{Task 3: Computational Time}\label{sec:experiment:time}

\textbf{Setup:}
We benchmark computational time per successful IK solution as a function of batch size (4--128), using the same test set as in Tasks~1. 
For each method, we initialize a batch of candidate seeds in parallel and record the time until the solver returns a solution that meets the Task~2 accuracy thresholds. We report results for: (i) vanilla \emph{Pink} and \emph{cuRobo}; (ii) \emph{IKFlow} seeding; and (iii) \name seeding. Reported time excludes scene loading and I/O. 
Since Pink is not natively parallel, we generate seeds in batches but run optimization sequentially.
For cuRobo, the maximum iteration budget is set to the saturation point observed in \cref{fig:time_per_solution} for each robot, while Pink uses 200 iterations as in Task~2.

\textbf{Metrics:}
The primary metric is \emph{time per solution} in milliseconds. We also study how computational time scales with the batch size, which controls the number of parallel seeds evaluated. Notably, \emph{Pink} is not a parallel IK solver; Pink evaluates them sequentially to obtain a precise solution.

\begin{figure*}[th!]
    \centering
    \includegraphics[width=1\linewidth]{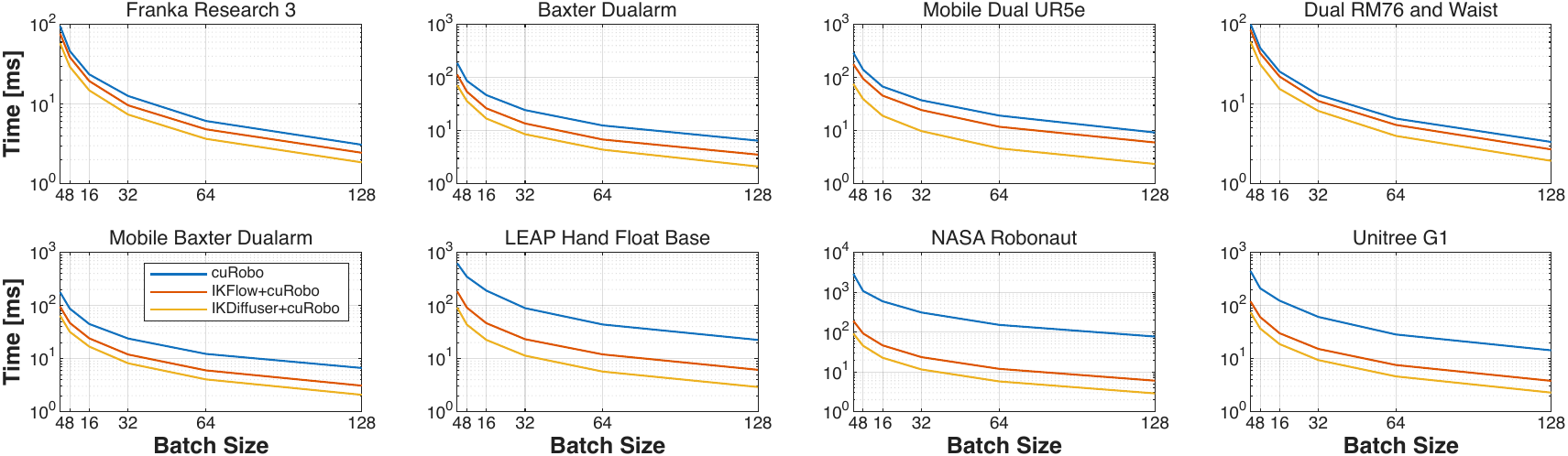}
    \caption{\textbf{Task 3: Computational Time.} Time-to-solution vs. batch size across eight robot platforms for cuRobo.}
    \label{fig:time_curobo}
\end{figure*}

\begin{figure*}[th!]
    \centering
    \includegraphics[width=1\linewidth]{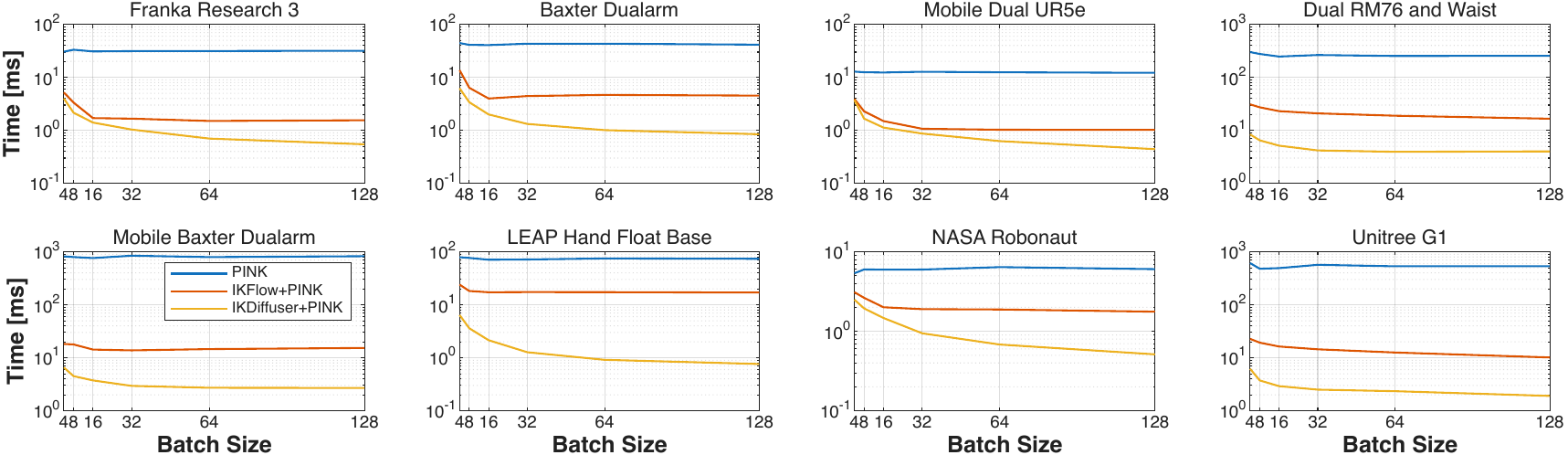}
    \caption{\textbf{Task 3: Computational Time.} Time-to-solution vs. batch size across eight robot platforms for Pink.}
    \label{fig:time_pink}
\end{figure*}

\textbf{Results and Analysis:}
Results are summarized in \cref{fig:time_pink,fig:time_curobo}.
Across all robots and both solvers, \name yields the lowest computational time and the most favorable scaling with batch size. In every platform, the \name seeded solver's curves lie below both vanilla and IKFlow-seeded baselines. The gap is largest on high-DoF and multi-arm systems (dexterous hand and humanoids), where \name typically cuts computational time by several factors relative to vanilla and remains faster than IKFlow across all batch sizes.

All methods benefit from a larger batch size, but \name exhibits the steepest drop, after which the gains taper. This indicates that a moderate number of diverse, high-quality seeds is sufficient for fast convergence; pushing batch size beyond 64 brings smaller incremental savings.
Without seeding, cuRobo often attains lower computational time than Pink at small batch size, but once seeded with \name, Pink rapidly becomes competitive or faster due to requiring far fewer iterations to converge (matching the iteration analysis in Task~2). On several platforms (\eg, Baxter Dualarm, Robonaut), \name+Pink reaches less than 1 millisecond solutions at batch size $\ge$ 32, whereas vanilla Pink remains orders of magnitude slower.


The computational time profile reflects the gains in Task~2: high-quality, diverse seeds from \name not only raise success rates but also \emph{decrease} time-to-solution, especially when combined with a moderate size of batching. Practically, this enables (i)~sub-$10$–$20$ milliseconds solutions on many platforms, (ii)~stable performance even on high-DoF systems, and (iii)~efficient scaling to large batched IK workloads for planning and control. In short, \name achieves near-linear latency reductions up to medium batch sizes and delivers the best overall time–accuracy trade-off across all robots and both solvers.

\begin{table*}[th!]
    \centering
    \caption{\textbf{Task 4:} Comparison of \name and Guided \name.}
    \resizebox{\textwidth}{!}{%
    \setlength{\tabcolsep}{4.5pt} 
    \setstretch{1.2} 
    \begin{tabular}{lcc >{\columncolor{red!20}}c >{\columncolor{red!20}}c >{\columncolor{red!20}}c >{\columncolor{red!20}}c | >{\columncolor{blue!20}}c >{\columncolor{blue!20}}c >{\columncolor{blue!20}}c >{\columncolor{blue!20}}c}
    \toprule
    \multicolumn{11}{c}{\textbf{Task 4: Objective-Guided IK Solution Generation (Warm-started)}} \\
    \toprule
    \multicolumn{3}{c}{\textbf{Robot Specifications}} 
    & \multicolumn{4}{>{\columncolor{red!20}}c}{\textbf{\name}} 
    & \multicolumn{4}{>{\columncolor{blue!20}}c}{\textbf{Warm-started \name}} \\
    \cmidrule(lr){1-3} \cmidrule(lr){4-7} \cmidrule(lr){8-11}
    Robot Platform & \textbf{$N_\text{ee}$} & DoF 
    & Coll. (\%)$\downarrow$ & Position (mm)$\downarrow$ & Angular (deg)$\downarrow$ & J.Diff. (\%)$\downarrow$ 
    & Coll. (\%)$\downarrow$ & Position (mm)$\downarrow$ & Angular (deg)$\downarrow$ & J.Diff. (\%)$\downarrow$ \\
    \midrule
    Franka Research 3     & 1 & 7  & 0.60 & 5.16 $\pm$ 3.86 & 1.09 $\pm$ 0.78 & 23.13 & 0.60 & 5.78 $\pm$ 3.79 & 1.26 $\pm$ 0.96 & 3.44 \\
    Baxter Dualarm        & 2 & 14 & 1.96 & 5.49 $\pm$ 2.50 & 1.26 $\pm$ 0.66 & 24.27 & 1.56 & 5.92 $\pm$ 2.60 & 1.37 $\pm$ 0.63 & 4.64 \\
    Mobile Dual UR5e      & 2 & 15 & 8.38 & 9.48 $\pm$ 4.75 & 1.49 $\pm$ 0.76 & 24.46 & 9.03 & 9.89 $\pm$ 3.94 & 1.56 $\pm$ 0.65 & 5.27 \\
    Dual RM76 and Waist   & 2 & 17 & 0.64 & 4.66 $\pm$ 1.76 & 1.63 $\pm$ 0.96 & 29.57 & 0.89 & 5.13 $\pm$ 2.04 & 1.89 $\pm$ 1.03 & 6.57 \\
    Mobile Baxter Dualarm & 2 & 17 & 3.78 & 9.49 $\pm$ 3.84 & 1.65 $\pm$ 0.78 & 23.92 & 3.97 & 9.86 $\pm$ 3.77 & 1.74 $\pm$ 0.77 & 7.11 \\
    LEAP Hand Float Base  & 4 & 22 & 5.02 & 1.79 $\pm$ 0.72 & 0.96 $\pm$ 0.34 & 6.33  & 4.84 & 1.82 $\pm$ 0.79 & 0.99 $\pm$ 0.36 & 1.46 \\
    NASA Robonaut         & 4 & 29 & 3.47 & 9.58 $\pm$ 4.55 & 1.56 $\pm$ 0.83 & 18.71 & 3.25 & 9.90 $\pm$ 4.71 & 1.62 $\pm$ 0.81 & 6.91 \\
    Unitree G1            & 4 & 29 & 5.97 & 5.37 $\pm$ 1.62 & 1.49 $\pm$ 0.48 & 14.98 & 5.38 & 5.57 $\pm$ 1.61 & 1.54 $\pm$ 0.47 & 4.95 \\
    \bottomrule
    \end{tabular}%
    }%
    \label{tab:warm_started}
\end{table*}

\begin{table*}[th!]
    \centering
    \caption{\textbf{Task 4:} Comparison of \name and Manipulability-aware \name.}
    \resizebox{\textwidth}{!}{%
    \setlength{\tabcolsep}{4.5pt} 
    \setstretch{1.2} 
    \begin{tabular}{lcc >{\columncolor{red!20}}c >{\columncolor{red!20}}c >{\columncolor{red!20}}c | >{\columncolor{blue!20}}c >{\columncolor{blue!20}}c >{\columncolor{blue!20}}c >{\columncolor{blue!20}}c}
    \toprule
    \multicolumn{10}{c}{\textbf{Task 4: Objective-Guided IK Solution Generation (Manipulability-aware)}} \\
    \toprule
    \multicolumn{3}{c}{\textbf{Robot Specifications}} 
    & \multicolumn{3}{>{\columncolor{red!20}}c}{\textbf{\name}} 
    & \multicolumn{4}{>{\columncolor{blue!20}}c}{\textbf{Manipulability-aware \name}} \\
    \cmidrule(lr){1-3} \cmidrule(lr){4-6} \cmidrule(lr){7-10}
    Robot Platform & \textbf{$N_\text{ee}$} & DoF 
    & Coll. (\%)$\downarrow$ & Position (mm)$\downarrow$ & Angular (deg)$\downarrow$ 
    & Coll. (\%)$\downarrow$ & Position (mm)$\downarrow$ & Angular (deg)$\downarrow$ & M.Imp. (\%)$\uparrow$ \\
    \midrule
    Franka Research 3 & 1 & 7  & 0.49 & 4.90 $\pm$ 2.84 & 1.05 $\pm$ 0.62 & 0.83 & 5.36 $\pm$ 2.93 & 1.19 $\pm$ 0.72 & 67.50 \\
    Baxter Dualarm & 2 & 14 & 1.78 & 5.25 $\pm$ 2.09 & 1.22 $\pm$ 0.51 & 2.18 & 6.03 $\pm$ 2.25 & 1.35 $\pm$ 0.62 & 26.22 \\
    Mobile Dual UR5e & 2 & 15 & 11.18 & 9.45 $\pm$ 4.16 & 1.48 $\pm$ 0.68 & 11.64 & 9.48 $\pm$ 4.43 & 1.49 $\pm$ 0.74 & 72.55 \\
    Dual RM76 and Waist & 2 & 17 & 0.66 & 4.46 $\pm$ 1.74 & 1.59 $\pm$ 0.83 & 1.07 & 4.84 $\pm$ 2.01 & 1.71 $\pm$ 1.07 & 73.23 \\
    Mobile Baxter Dualarm & 2 & 17 & 3.72 & 9.44 $\pm$ 3.66 & 1.64 $\pm$ 0.73 & 4.77 & 9.84 $\pm$ 4.59 & 1.91 $\pm$ 1.02 & 38.38 \\
    LEAP Hand Float Base & 4 & 22 & 2.80 & 1.73 $\pm$ 0.82 & 0.93 $\pm$ 0.49 & 3.27 & 2.32 $\pm$ 0.95 & 1.19 $\pm$ 0.51 & 0.43 \\
    NASA Robonaut & 4 & 29 & 3.20 & 9.28 $\pm$ 3.65 & 1.51 $\pm$ 0.57 & 4.79 & 9.93 $\pm$ 4.25 & 1.72 $\pm$ 0.98 & 15.88 \\
    Unitree G1 & 4 & 29 & 6.12 & 5.32 $\pm$ 1.69 & 1.48 $\pm$ 0.59 & 8.28 & 5.99 $\pm$ 2.23 & 1.74 $\pm$ 1.08 & 24.69 \\
    \bottomrule
    \end{tabular}%
    }%
    \label{tab:manip_aware}
\end{table*}

\subsection{Task 4: Objective-Guided IK Solution Generation}\label{sec:experiment:objective}

\textbf{Setup:} 
To assess the adaptability of \name to task-specific objectives, we investigate two guided variants: (i) warm-started \name, and (ii) manipulability-aware \name. Both variants build directly upon the standard experimental setup introduced in~\cref{sec:experiment:ik_gen}, with modifications reflecting their respective objectives. In the warm-started configuration, the target end-effector pose is perturbed within a $2$~cm radius of the nominal target, while the solver is initialized with the prior joint configuration. This setup evaluates the ability of the model to exploit local information and maintain trajectory continuity. For the manipulability-aware configuration, the canonical generation pipeline is preserved, but the objective is augmented with a manipulability-based term that explicitly encourages configurations yielding higher dexterity. 

\textbf{Metrics:} 
In addition to the collision rate, end-effector position error and angular error, we introduce two task-oriented measures. For the warm-started case, we define Joint Difference (J.Diff.) as the normalized mean deviation from the initial joint configuration, which directly quantifies motion smoothness and adherence to prior states. For the manipulability-aware case, we measure the Manipulability Improvement Ratio (M.Imp.), defined as the relative percentage increase in manipulability compared to the standard \name. 

\textbf{Results and Analysis:} 
The results summarized in \cref{tab:warm_started} indicate that warm-started \name markedly reduces joint differences across all tested platforms, achieving up to $80\%$ reduction in some high-DoF manipulators, while maintaining end-effector accuracy and collision avoidance similar to the standard model. This property is important in applications such as teleoperation or reactive control, where solutions that remain close to the prior configuration ensure smoother and more stable execution without sacrificing precision.  

Similarly, as reported in \cref{tab:manip_aware}, manipulability-aware \name consistently improves manipulability across diverse robot morphologies, with the most significant gains observed in highly redundant systems such as dual-arm and humanoid platforms. Importantly, these improvements are achieved with only marginal trade-offs in position accuracy ($<1$~mm on average) and negligible increases in collision rate, both of which can be readily corrected by a lightweight deterministic IK refinement step.  

Together, these findings highlight the inherent flexibility of \name. Without any retraining, the model can be seamlessly adapted to new task objectives---whether prioritizing motion smoothness through warm-start guidance or enhancing dexterity via manipulability objectives. This adaptability underscores its potential as a general-purpose generative IK solver, suitable for a wide spectrum of real-world scenarios where smooth execution, safety, and dexterity are all required.

\begin{figure*}[th!]
    \centering
    \includegraphics[width=\linewidth]{figures/mar_inf/marginal_inference_3_6.pdf}
    \caption{\textbf{Task 5:} Illustration of marginal inference results. Joints belonging to chains with unconstrained end-effectors remain fixed.}
    \label{fig:marginal_inference}
\end{figure*}

\subsection{Task 5: Marginal Inference} \label{sec:experiment:marginal}

\textbf{Setup:} We evaluate \name under \emph{partially specified} \ac{ik} queries, where only a subset of end-effectors are assigned target poses. 
This setting investigates the model's marginal inference capability: satisfying the specified goals while implicitly marginalizing over the unobserved end-effectors. 
For a robot with $N_\text{ee}$ end-effectors, we randomly mask a subset and provide goals for the remaining ones. 
We uniformly sample the number of masked end-effectors $m \in \{0,\dots,N_\text{ee}-1\}$ (ensuring at least one end-effector remains unmasked), and then uniformly sample which $m$ end-effectors are masked. 
Masked end-effectors are replaced by the empty token in the conditioning sequence, while the unmasked end-effectors are conditioned on their target poses, forming the mixed conditioning input as described in \cref{eqn:mixed_conditioning}. 
We evaluate \name on three representative platforms (dual-arm, humanoid, and dexterous hand) with varying numbers of end-effectors.

\textbf{Metrics:} We follow the protocol in Task~1, but compute pose errors \emph{only on the unmasked end-effectors} (\ie, those with specified goals). 
We report collision rate (Coll.~\%), Cartesian position error (mm), and angular error (deg) on the goal-specified end-effectors to assess whether \name can meet the provided goals while marginalizing the masked ones.

\begin{table}[t!]
    \centering
    \caption{\textbf{Task 5}: \name on partially specified IK queries.}
    \resizebox{\columnwidth}{!}{%
    \setlength{\tabcolsep}{2pt} 
    \setstretch{1.2} 
    \begin{tabular}{lccccc}
    \toprule
    \multicolumn{6}{c}{\textbf{Task 5: Marginal Inference}} \\
    \toprule
    \multicolumn{3}{c}{\textbf{Robot Specification}}
    & \multicolumn{3}{c}{\textbf{\name}} \\
    \cmidrule(lr){1-3} \cmidrule(lr){4-6}
    Robot Platform & \textbf{$N_\text{ee}$} & DoF
    & Coll. (\%)$\downarrow$ & Position (mm)$\downarrow$ & Angular (deg)$\downarrow$ \\
    \midrule
    Dual RM76 and Waist     & 2 & 17 & 1.04 & 5.61 $\pm$ 1.77 & 1.74 $\pm$ 0.87 \\
    LEAP Hand Float Base    & 4 & 22 & 3.03 & 2.11 $\pm$ 0.84 & 1.18 $\pm$ 0.45 \\
    Unitree G1              & 4 & 29 & 7.18 & 6.06 $\pm$ 1.83 & 1.63 $\pm$ 0.59 \\
    \bottomrule
    \end{tabular}%
    }%
    \label{tab:marginal_inference}
\end{table}

\textbf{Results and Analysis:} \cref{tab:marginal_inference} summarizes marginal inference performance on the three robots, and \cref{fig:marginal_inference} provides an illustration.
Overall, \name remains robust, achieving low pose errors across all robots. 
As the number of masked end-effectors $m$ is uniformly sampled, the test cases span varying degrees of under-constraint: larger $m$ yields a less constrained posterior with a broader feasible solution set, increasing redundancy and potentially amplifying inter-branch couplings. 
Despite this, \name maintains stable performance with low variance across runs (this variance is comparable to fully specified \ac{ik} queries in Task~1), suggesting that it does not depend on a fixed and fully specified conditioning pattern and can instead leverage the learned configuration distribution to infer compatible solutions. 
Compared to the setting in Task~1 (see \cref{tab:ik_quality}), pose errors on the specified end-effectors increase only moderately (typically $<1$~mm), which is expected since the removal of conditioning information enlarges the feasible solution set and introduces additional redundancy. 
Overall, these results validate \name's marginal inference capability: it enforces observed end-effector goals while implicitly integrating over missing constraints, enabling a unified model to handle both fully specified and partially specified \ac{ik} queries without retraining the model.


\subsection{Task 6: Error Analysis}\label{sec:experiment:error}

While the previous experiments demonstrate the general efficacy of \name, it is crucial to understand how specific kinematic properties, \ie, the physical scale of the robot and the dimensionality of the chain, contribute to pose errors. To isolate these factors from the complexities of specific robot morphologies (\eg, joint limits or offsets), we conducted a controlled ablation study using synthetic serial chains.

\textbf{Setup}:
We constructed two sets of 2D serial manipulators (without self-collisions) to test two hypotheses:
1)~We fixed the chain complexity at 3 \acp{dof} but varied the link lengths ($0.3$\,m, $0.6$\,m, $0.9$\,m), resulting in total chain lengths ranging from $0.9$\,m to $2.7$\,m. This setup allows us to investigate whether position error scales linearly with robot size (geometric error propagation) or if the model struggles with larger task spaces.
2)~We fixed the total chain length at $0.9$\,m but increased the number of joints from 3 to 5 (adjusting individual link lengths accordingly). This isolates the effect of dimensionality, testing the model's ability to resolve redundancy and accumulate joint rotations within a fixed workspace volume.

\textbf{Metric}: We report positional and angular errors, consistent with the evaluation metrics used in previous tasks. 

\begin{table}[th!]
    \centering
    \caption{\textbf{Task 6:} Impact of robot configuration on \name.}
    \resizebox{\columnwidth}{!}{%
    \setlength{\tabcolsep}{4.5pt} 
    \setstretch{1.2} 
    \begin{tabular}{ccc|cc}
    \toprule
    \multicolumn{5}{c}{\textbf{Task 6: Error Analysis}} \\
    \toprule
    \multicolumn{3}{c}{\textbf{Robot Configuration}} 
    & \multicolumn{2}{c}{\textbf{\name}} \\
    \cmidrule(lr){1-3} \cmidrule(lr){4-5}
    DoF & Link Len. (m) & Total Len. (m)  
    & Position (mm)$\downarrow$ & Angular (deg)$\downarrow$ \\ 
    \midrule
    3 & 0.3 & 0.9   & 4.18 $\pm$ 1.21 & 0.46 $\pm$ 0.13 \\
    3 & 0.6 & 1.8   & 6.82 $\pm$ 2.47 & 0.48 $\pm$ 0.12 \\
    3 & 0.9 & 2.7   & 10.08 $\pm$ 4.38 & 0.49 $\pm$ 0.16 \\
    \midrule
    3 & 0.3 & 0.9   & 4.18 $\pm$ 1.21 & 0.46 $\pm$ 0.13 \\
    4 & 0.225 & 0.9 & 5.01 $\pm$ 1.33 & 0.56 $\pm$ 0.15 \\
    5 & 0.18 & 0.9 & 5.33 $\pm$ 1.26 & 0.64 $\pm$ 0.19 \\
    \bottomrule
    \end{tabular}%
    }%
    \label{tab:error_analysis}
\end{table}

\textbf{Results and Analysis:}
The results, summarized in \cref{tab:error_analysis}, reveal two distinct error mechanisms:
In the variable scale experiments (Rows 1--3), we observe that the angular error remains stable ($\approx 0.46^{\circ}$ to $0.49^{\circ}$) regardless of link length. This indicates that \name predicts joint configurations with consistent precision, independent of the robot's physical size. However, the Cartesian position error increases significantly, from $4.18$\,mm to $10.08$\,mm. This behavior is consistent with the geometric lever-arm effect, where a constant angular deviation at the root joint results in a larger arc-length error at the end-effector as the link length increases. Thus, the degradation in position accuracy for larger robots is primarily a geometric consequence rather than a failure of the learning model itself.
In the variable complexity experiments (Rows 4--6), extending the chain from 3 to 5 DoFs (while keeping total length constant) leads to a degradation in both position ($4.18$ vs. $5.33$\,mm) and angular ($0.46^{\circ}$ vs. $0.64^{\circ}$) accuracy. Unlike the scale experiment, here the angular error increases by nearly 40\%. This suggests that as the dimensionality of the configuration space increases, the conditional distribution $p(\boldsymbol{q}|\mathcal{X})$ becomes more complex to model, and the accumulation of small errors across more joints makes the inverse mapping more challenging. Nevertheless, the increase in position error is relatively small, demonstrating that \name maintains reasonable robustness even as kinematic redundancy increases.


\setstretch{1.}

\section{Discussion}\label{sec:discussion}

\subsection{Key Findings}
\textbf{Scalability and Adaptability:}
First, our results demonstrate that formulating IK as a generative task through our structure-agnostic formulation offers superior scalability and adaptability compared to traditional regression or mapping approaches. Existing learning-based solvers typically rely on specific architectures that couple the input dimension to a fixed number of end-effectors and incorporate specific task constraints during training. In contrast, our proposed tokenized formulation decouples the robot structure from the network architecture, enabling \name to naturally scale to varying numbers of end-effectors and generalize robustly across arbitrary kinematic trees and unseen pose configurations.
Furthermore, this probabilistic formulation unlocks adaptability at the inference phase that regression models lack. By separating the learned kinematic prior from task-specific constraints, \name can handle partially specified end-effector goals (via masked marginalization) and incorporate novel objectives---as long as the objective is differentiable---solely through gradient guidance during sampling. This allows the solver to adapt to changing task requirements on the fly without the computational cost and engineering burden of retraining the model for every new constraint or task variant.

\textbf{The Complementary Nature of Learning and Optimization:} Second, we observe a distinct trade-off: while learning-based models provide excellent global structure, they inherently suffer from precision limitations due to their statistical nature. Diffusion models approximate the data distribution, which rarely yields the exact numerical precision ($< 10^{-5}$ errors) required for high-tolerance industrial tasks. Conversely, numerical optimization solvers offer high precision but are locally optimal and highly sensitive to initialization. 

Our experiments reveal that these two paradigms are highly complementary. \name acts as a robust ``global search'' mechanism that identifies valid basins of attraction, while the optimization solver acts as a ``local refiner''. This synergy is evidenced by the results in \cref{sec:experiment:seed}, where seeding cuRobo with \name increased the success rate on the Unitree G1 from 21.01\% to 96.96\%. This complementary relationship mirrors successful paradigms in other domains, such as protein folding (where neural networks predict structures that are physically refined~\cite{jumper2021highly}) and motion planning (where learned priors warm-start trajectory optimizers~\cite{huang2025diffusionseeder}). \name establishes that this hybrid ``Generate-then-Optimize'' pipeline is the most viable path for fast and constraint planning of complex kinematic trees.

\textbf{Data Efficiency in Model Training:}
Third, our results indicate that \name does not require an exhaustive dataset even as robot complexity (e.g., DoF and branching) increases.
While learning a direct mapping with strong in-distribution guarantees usually require near-complete domain coverage, this quickly becomes intractable: for a 29-DoF humanoid with each joint spanning $[-180^\circ, 180^\circ]$, a naive $1^\circ$ discretization implies $360^{29} \approx 1.35\times 10^{74}$ configurations.
In contrast, training on $10$M samples ($10^8$) per platform already yields millimeter-level accuracy in position ($\sim 6$mm) and rotation ($\sim 1.7^\circ$) on high-DoF humanoids, suggesting that accurate \ac{ik} solutions can be learned from a vanishingly small fraction of the combinatorial joint grid.

We also find that reducing the training set to $1$M samples can still produce decent performance on less-complex morphologies, such as serial chains and dual-arm systems, but does not reach the precision achieved with larger datasets on more complex platforms (\eg, humanoids), where coupled constraints and high-dimensional redundancy amplify data demands.
We do not exhaustively sweep the size of the data set to chase marginal gains, as 10 million samples already provide strong and stable performance across various robots.
In practice, the scale of the dataset can be treated as a tunable knob: one may start with a moderate budget for simpler systems and increase or specialize data collection when higher accuracy or robustness under complex kinematic coupling is required.

\subsection{Future Directions}

\textbf{Architectural Exploration:} In this work, our primary focus was to validate the efficacy of the structure-agnostic formulation and the diffusion-based probabilistic inference. We did not exhaustively optimize the underlying neural network architecture (\eg, exploring different Transformer variants, Mamba backbones, or varying attention mechanisms). Although the current architecture delivers substantial improvements over state-of-the-art baselines, future work could follow our probabilistic formulation and explore more parameter-efficient or inference-optimized backbones to further reduce computational latency without compromising generation quality.


\textbf{Behavior at Workspace Boundaries:} A central challenge in learning-based IK is handling targets outside the robot's reachable workspace. Such targets are entirely out-of-distribution because the training data contains no unreachable cases. As detailed in our supplementary analysis (see \cref{sec:appendix:ood} and \cref{fig:ik_unreachable}), \name exhibits a desirable ``graceful degradation'' in these regimes. Similar to pure optimization solvers, \name consistently proposes valid, ``closest-feasible'' configurations. When used to seed optimization, these proposals reliably stabilize the solver, ensuring it converges to the best possible solution even when the exact target is unreachable. Future work could explicitly model this ``reachability probability'' to provide a confidence score alongside the generated solution.

\textbf{Integration with Downstream Tasks:} Finally, the speed and diversity of \name position it as a powerful primitive for higher-level planning frameworks. In multi-arm motion planning under end-effector constraints, \name can serve as a high-frequency goal sampler that rapidly populates the configuration space with valid start/goal pairs, even in cluttered environments. Furthermore, in \ac{tamp}~\cite{jiao2021efficient,garrett2020integrated}, where geometric feasibility checks often form a computational bottleneck, \name can function as a fast, batch-parallelized feasibility oracle. By rapidly filtering infeasible high-level actions before invoking costly motion planners, \name has the potential to significantly accelerate the planning horizon for general-purpose robots.


\begin{table*}[th!]
    \centering
    \caption{Comparison of IKDiffuser w/o and w/ sequentializing end-effector poses.}
    \resizebox{\textwidth}{!}{%
    \setlength{\tabcolsep}{6pt} 
    \setstretch{1.2} 
    \begin{tabular}{lcc >{\columncolor{red!20}}c >{\columncolor{red!20}}c >{\columncolor{red!20}}c >{\columncolor{blue!20}}c >{\columncolor{blue!20}}c >{\columncolor{blue!20}}c}
    \toprule
    \multicolumn{9}{c}{\textbf{Appendix: Ablation Studies for structure-agnostic Conditioning}} \\
    \toprule
    \multicolumn{3}{c}{\textbf{Robot Specification}} & \multicolumn{3}{>{\columncolor{red!20}}c}{\textbf{w/o sequentializing end-effector poses}} & \multicolumn{3}{>{\columncolor{blue!20}}c}{\textbf{w/ sequentializing end-effector poses}} \\
    \cmidrule(lr){1-3} \cmidrule(lr){4-6} \cmidrule(lr){7-9}
    Robot Platform & \textbf{$N_\text{ee}$} & DoF & Collision (\%)$\downarrow$ & Position (mm)$\downarrow$ & Angular (deg)$\downarrow$ & Collision (\%)$\downarrow$ & Position (mm)$\downarrow$ & Angular (deg)$\downarrow$ \\
    \midrule
    Franka Research 3      & 1 & 7  & 0.51 & 5.12 $\pm$ 2.91 & 1.25 $\pm$ 0.71 & \textbf{0.49} & \textbf{4.90 $\pm$ 2.84} & \textbf{1.05 $\pm$ 0.62} \\
    Baxter Dualarm          & 2 & 14 & 3.23 & 9.87 $\pm$ 3.63 & 2.52 $\pm$ 1.22 & \textbf{1.78} & \textbf{5.25 $\pm$ 2.09} & \textbf{1.22 $\pm$ 0.51} \\
    Dual RM76 with Waist     & 2 & 17 & 2.04 & 9.12 $\pm$ 3.21 & 2.31 $\pm$ 1.07 & \textbf{0.66} & \textbf{4.46 $\pm$ 1.74} & \textbf{1.59 $\pm$ 0.83} \\
    Mobile Dual UR5e        & 2 & 15 & 18.76 & 18.71 $\pm$ 7.81 & 3.51 $\pm$ 1.32 & \textbf{8.18} & \textbf{9.45 $\pm$ 4.16} & \textbf{1.48 $\pm$ 0.68} \\
    Mobile Baxter Dualarm   & 2 & 17 & 7.41 & 17.86 $\pm$ 6.93 & 3.26 $\pm$ 1.12 & \textbf{3.72} & \textbf{9.44 $\pm$ 3.66} & \textbf{1.64 $\pm$ 0.73} \\
    LEAP Hand Float Base    & 4 & 22 & 8.27 & 8.86 $\pm$ 4.03 & 5.21 $\pm$ 1.87 & \textbf{2.85} & \textbf{1.73} $\pm$ \textbf{0.72} & \textbf{0.93} $\pm$ \textbf{0.33} \\
    NASA Robonaut           & 4 & 29 & 9.16 & 29.37 $\pm$ 9.67 & 4.72 $\pm$ 1.97 & \textbf{3.20} & \textbf{9.28 $\pm$ 3.65} & \textbf{1.51 $\pm$ 0.57} \\
    Unitree G1              & 4 & 29 & 22.23 & 19.65 $\pm$ 5.93 & 6.43 $\pm$ 2.23 & \textbf{6.35} & \textbf{5.31} $\pm$ \textbf{1.56} & \textbf{1.48} $\pm$ \textbf{0.49} \\
    \bottomrule
    \end{tabular}%
    }%
    \label{tab:ablation}
\end{table*}

\section{Conclusion}\label{sec:conclusion}

In this work, we introduce \name, a unified, scalable generative framework for solving the Inverse Kinematics (IK) problem for arbitrary kinematic trees. 
By formulating IK as a probabilistic inference task within a diffusion model, we address the longstanding challenges of scalability to high-dimensional configuration spaces and complexity of inter-branch constraints that traditional optimization and regression-based approaches suffer from. 
Our structure-agnostic formulation enables a unified model architecture to generalize across varied robot morphologies---from serial manipulators to complex humanoids and dexterous hands---without requiring robot-specific customization. 
Extensive evaluations on eight robotic platforms demonstrate that \name significantly outperforms state-of-the-art baselines in terms of solution diversity, accuracy, and self-collision avoidance. Through masked marginalization and objective-guided sampling, the model is capable of handling partially constrained end-effector goals and task-specific requirements without further retraining. 
Furthermore, we demonstrated the critical role of generative priors; using \name to seed optimization-based solvers boosted success rates on high-DoF systems while maintaining millisecond-level computational time.
As robotic systems evolve toward greater complexity and autonomy, we envision \name stands as an essential foundation, empowering general-purpose robots to interact with the physical world with a level of adaptability and robustness previously unattainable.

\appendix

\subsection{Ablation Studies for structure-agnostic Conditioning} \label{subsec:ablation}

To validate the design choices behind \name, we conducted an ablation study to assess the impact of our structure-agnostic conditioning formulation. Specifically, we investigate whether representing end-effector poses as a sequence of independent tokens offers an advantage over the traditional approach of concatenating them into a flat vector.

\textbf{Setup:} We compare two model variants:
\begin{itemize}[leftmargin=*]
    \item \textbf{w/ Sequentializing (Proposed):} The standard \name architecture, where each of the $N_\text{ee}$ end-effector poses is embedded into a distinct token. This allows the cross-attention mechanism to model relationships between specific end-effectors and the joint configuration tokens independently.
    \item \textbf{w/o Sequentializing (Baseline):} A variant where all $N_\text{ee}$ end-effector poses are concatenated into a single flat vector of dimension $N_\text{ee} \times 7$. This vector is projected into a single embedding token conditioned on the diffusion process. This approach mimics the conditioning strategy used in prior work on dual-arm systems and standard conditional diffusion models.
\end{itemize}
Both variants were trained using identical hyperparameters, dataset sizes (10M samples), and training duration (200 epochs) to ensure a fair comparison.

\textbf{Metrics:} We utilize the same evaluation metrics defined in \cref{sec:experiment:ik_gen}: \textit{Collision Rate} (\%), \textit{Position Error} (mm), and \textit{Angular Error} (deg). Lower values indicate better performance for all metrics.

\textbf{Results and Analysis:} The quantitative results are summarized in \cref{tab:ablation}. 
For simple serial chains like the \textit{Franka Research 3}, the performance difference between the two methods is marginal (\eg, 0.49\% vs. 0.51\% collision rate). This is expected, as single-arm systems lack inter-branch constraints, making the flattened representation sufficient to capture the kinematic constraints.

However, the advantages of our formulation become more obvious as the kinematic structure becomes more complex. On multi-arm and high-DoF systems, the baseline exhibits a sharp degradation in performance. For instance, on the \textit{Mobile Dual UR5e}, the collision rate spikes to 18.76\% with the baseline, compared to 8.18\% with \name. Similarly, on the \textit{NASA Robonaut}, the baseline produces significantly higher position errors ($29.37$ mm vs. $9.28$ mm) and nearly triple the collision rate ($9.16\%$ vs. $3.20\%$).

This performance gap highlights that flattening end-effector poses obscures the structural topology of the kinematic trees. By sequentializing the inputs, \name enables the attention mechanism to explicitly reason about the coupling between specific end-effectors and their shared joints (\eg, a waist or mobile base). This structural awareness is critical for resolving the complex inter-chain constraints and self-collision constraints inherent in kinematic trees.


\begin{figure*}[th!]
    \centering
    \begin{subfigure}[b]{\linewidth}
    \centering
    \includegraphics[width=\linewidth, trim={0 0 0 0},clip]{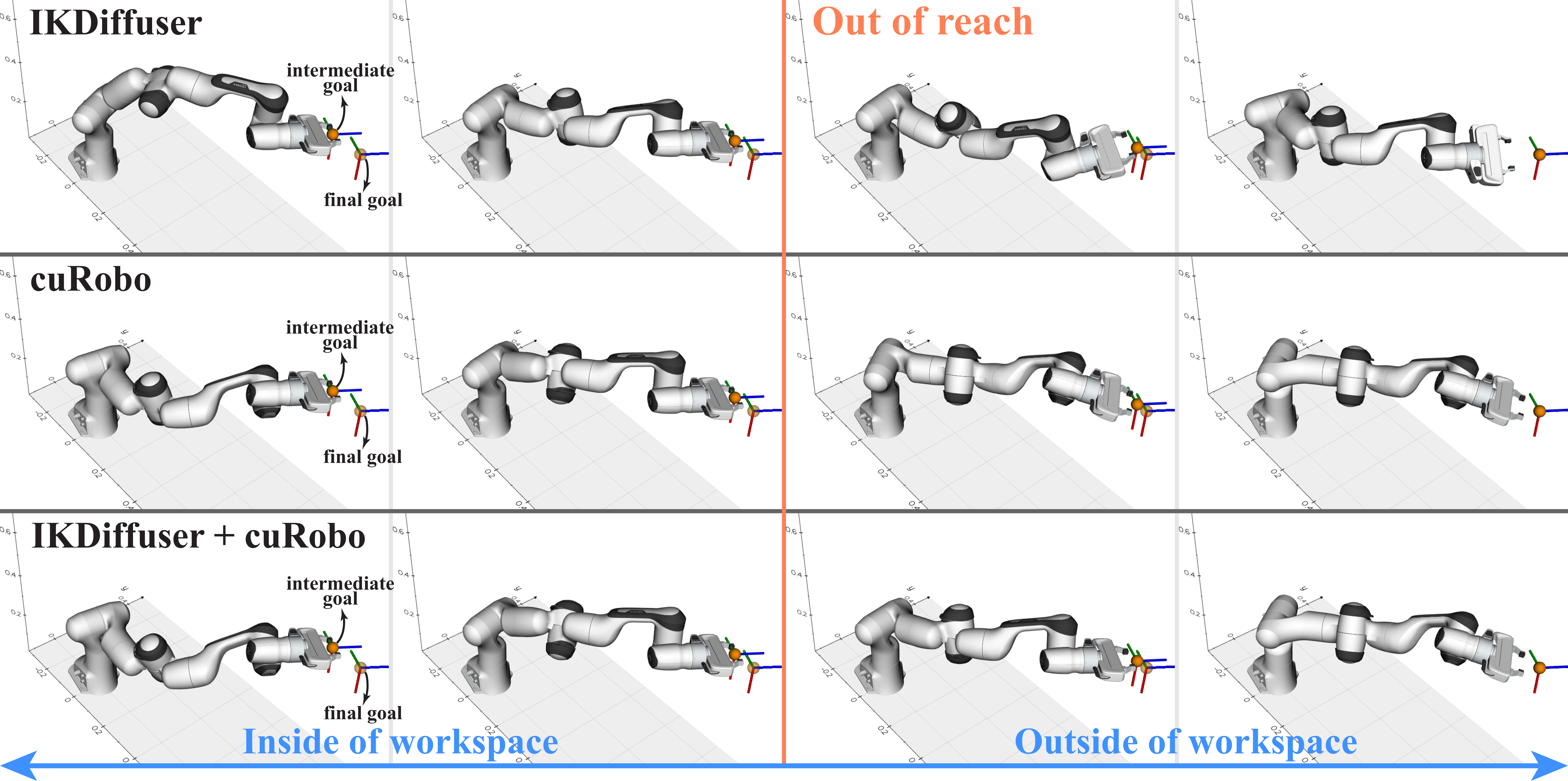}
        \caption{Franka Research 3}
        \label{fig:unreachable_franka}
    \end{subfigure}
    \\
    \begin{subfigure}[b]{\linewidth}
        \centering
        \includegraphics[width=\linewidth, trim={0 0 0 0},clip]{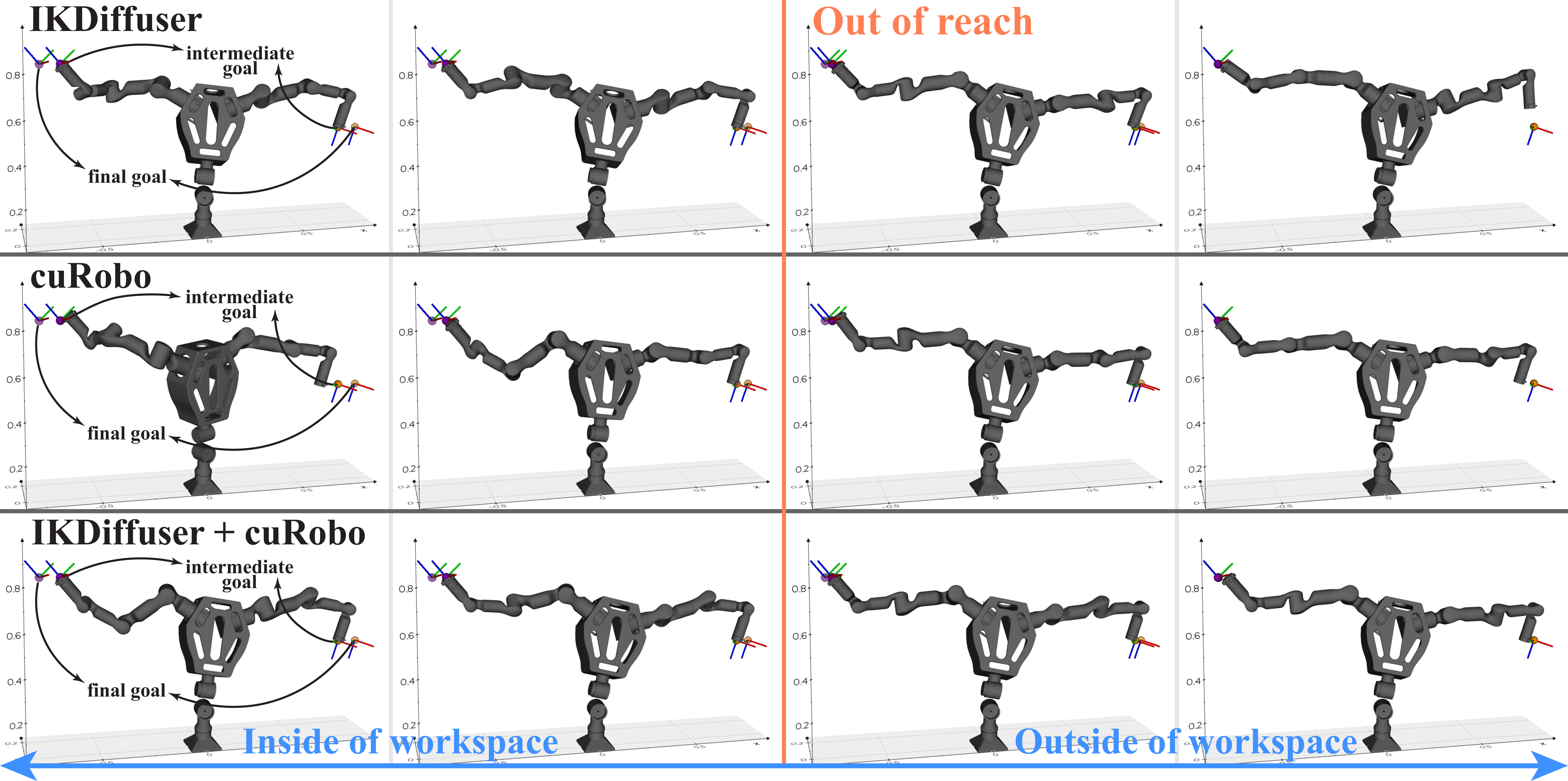}
        \caption{Dual RM76 with Waist}
        \label{fig:unreachable_tong_dualarm}
    \end{subfigure}
    \caption{\textbf{IK solutions for unreachable end-effector goals.} Rows show solutions produced by each method as the intermediate goal (solid axes) moves progressively (5cm each move) to the final goal (semi-transparent axes), which is \emph{far outside} the reachable workspace.}
    \label{fig:ik_unreachable}
\end{figure*}

\subsection{Transition to Unreachable End-Effector Goals}\label{sec:appendix:ood}
Beyond standard success rates and mean errors, we seek to \emph{visually characterize failure modes} of IK solvers when the target end-effector pose is \emph{outside the reachable workspace}. This presents a regime that often arises in manipulation near workspace boundaries. This qualitative study asks: (i) do solvers degrade gracefully by returning the closest-feasible pose, (ii) does a generative prior (\name) provide informative proposals near the boundary, and (iii) can such proposals reliably \emph{stabilize} a downstream optimizer (cuRobo) on challenging, redundant systems? We evaluate three policies: \name, cuRobo, and cuRobo seeded with \name, on two platforms: a single-arm Franka manipulator and a dual-arm manipulator with a waist. 
\cref{fig:ik_unreachable} shows a qualitative result, for each robot, we place intermediate end-effector frames progressively beyond the kinematic boundary by translating from a reachable pose along a fixed direction (5cm each move), producing a sweep from ``just inside'' to ``far outside''. 
For each target, the solver returns its best feasible configuration, and we visualize the intermediate end-effector goal frame for comparison against the final goal frame.

For targets outside the kinematic reach, our experiments (\cref{fig:ik_unreachable}) reveal three consistent behaviors. First, on the single-arm Franka, all methods recover a valid configuration when the goal lies inside the reachable set; once the goal moves beyond it, the optimization baseline (cuRobo) converges to the nearest feasible posture in task space, while \name also approaches the goal but exhibits a monotonic increase in residual pose error as the goal recedes from the boundary. Second, in the more challenging dual–arm setting, plain cuRobo struggles to obtain precise solutions even for targets nominally inside reach, reflecting sensitivity to nonconvexity and poor local minima. In contrast, \name reliably proposes solutions close to the target, indicating that its generative prior captures useful structure in the multi-arm manifold and exploits redundancy to reduce task space error. Third, combining the two yields the best of both worlds: cuRobo seeded with \name consistently converges to exact (or near-exact) solutions wherever they exist, demonstrating that diffusion-generated proposals substantially improve optimization robustness and precision in multi-arm IK, while still degrading gracefully (closest-feasible behavior) when targets are unreachable.

\bibliographystyle{ieeetr}
\bibliography{reference_header,reference}

\end{document}